\pdfoutput=1

\documentclass[11pt]{article}

\usepackage[final]{acl}

\usepackage{times}
\usepackage{latexsym}

\usepackage[T1]{fontenc}

\usepackage[utf8]{inputenc}

\usepackage{microtype}

\usepackage{inconsolata}

\usepackage{graphicx}

\usepackage{amsmath}
\usepackage{booktabs}
\usepackage{multirow}
\usepackage{array}
\usepackage{hyperref}
\usepackage{amsfonts}
\usepackage{amsthm}

\usepackage[T1]{fontenc}
\usepackage[utf8]{inputenc}
\usepackage{listings}
\usepackage{geometry}

\makeatletter
\newtheorem*{rep@theorem}{\rep@title}
\newcommand{\newreptheorem}[2]{%
\newenvironment{rep#1}[1]{%
 \def\rep@title{#2 \ref{##1}}%
 \begin{rep@theorem}}%
 {\end{rep@theorem}}}
\makeatother

\usepackage{thmtools,thm-restate}

\newreptheorem{theorem}{Theorem}

\newreptheorem{proposition}{Proposition}

\newreptheorem{lemma}{Lemma}

\newreptheorem{corollary}{Corollary}

\makeatletter
\newcommand{\thickhline}{%
    \noalign {\ifnum 0=`}\fi \hrule height 1pt
    \futurelet \reserved@a \@xhline
}
\newcolumntype{"}{@{\hskip\tabcolsep\vrule width 1pt\hskip\tabcolsep}}
\makeatother

\DeclareMathAlphabet\mathbfcal{OMS}{cmsy}{b}{n}

\newcommand{\methodA}{\textit{BCR}}
\newcommand{\methodB}{\textit{IPR}}

\usepackage[nolist,nohyperlinks]{acronym}
\makeatletter
\newcommand{\removelatexerror}{\let\@latex@error\@gobble}
\makeatother
\usepackage[font=small]{caption}
\usepackage{url}
\usepackage{wrapfig}
\usepackage{enumitem}

\title{Towards Improved Preference Optimization Pipeline: \\ from Data Generation to Budget-Controlled Regularization}

\author{
 \textbf{Zhuotong Chen\textsuperscript{1,2}},
 \textbf{Fang Liu\textsuperscript{1}},
 \textbf{Jennifer Zhu\textsuperscript{1}},
 \textbf{Wanyu Du\textsuperscript{1}},
 \textbf{Yanjun Qi\textsuperscript{1,2}}
\\
 \textsuperscript{1}AWS Bedrock Science
\\
 \small{
    \textsuperscript{2}
   \textbf{Correspondence:} \href{mailto:email@domain}{zhuotong@amazon.com},
   \href{mailto:email@domain}{yanjunqi@amazon.com}
 }
}

\begin{document}
\maketitle
\begin{abstract}
Direct Preference Optimization (DPO) and its variants have become the de facto standards for aligning large language models (LLMs) with human preferences or specific goals. 
However, DPO requires high-quality preference data and suffers from unstable preference optimization.
In this work, we aim to improve the preference optimization pipeline by taking a closer look at preference data generation and training regularization techniques. 
For preference data generation, we demonstrate that existing scoring-based reward models produce unsatisfactory preference data and perform poorly on out-of-distribution tasks. 
This significantly impacts the LLM alignment performance when using these data for preference tuning. 
To ensure high-quality preference data generation, we propose an iterative pairwise ranking mechanism that derives preference ranking of completions using pairwise comparison signals.
For training regularization,
we observe that preference optimization tends to achieve better convergence when the LLM predicted likelihood of preferred samples gets slightly reduced. 
However, the widely used supervised next-word prediction regularization strictly prevents any likelihood reduction of preferred samples. 
This observation motivates our design of a budget-controlled regularization formulation.
Empirically we show that combining the two designs leads to aligned models that surpass existing SOTA across two popular benchmarks.
\end{abstract}

\section{Introduction}
\label{sec: introduction}
Recently, Direct Preference Optimization (DPO) \citep{rafailov2024direct} and its variants \citep{meng2024simpo, azar2024general, ethayarajh2024kto, liu2024provably, pal2024smaug, xucontrastive} have gained popularity over traditional reinforcement learning from human feedback (RLHF) \citep{ziegler2019fine}, which involves training a reward model followed by reinforcement learning. 
DPO-based methods bypass the need for a reward model in optimization by directly optimizing the target model using preference data, leading to simpler and more efficient training.

The pipeline of DPO (and its variants) consists of two key stages: (1) collecting preference data by scoring various outputs generated by the target LLM model, and (2) performing direct optimization using the preference data. 

The first stage of constructing preference data involves two steps:
(1) the target model generates multiple completions for each input prompt;
(2) then a reward model selects preferred and dispreferred completions from these candidates for each prompt \citep{xiong2024iterative, meng2024simpo}.
Existing open-sourced reward models are mostly based on a classification architecture by modifying the last layer of a LLM \citep{skyworkreward2024, ArmoRM, wang2024arithmetic}.
This scoring-based approach for evaluating the quality of a prompt-completion pair introduces considerable noise \citep{cui2023ultrafeedback, ganguli2022red, guo2024direct}, and the issue becomes even more when the downstream task is out-of-distribution compared to the training data used to construct the reward model.

After constructing high-quality preference data, standard preference optimization algorithms compute the relative probability of selecting one completion over another by using pairs of preferred and dispreferred completions \citep{rafailov2024direct, meng2024simpo, azar2024general}. Optimizing towards this relative objective can potentially lead to a reduction of target model's predicted likelihood of the preferred completions, as long as the relative probability between the preferred and dispreferred completions increases with the preference optimization. 
This may cause training instability issue \citep{pal2024smaug, feng2024towards, liu2024provably}.
To address the challenge, several regularization techniques have been proposed to utilize supervised next-word prediction of the preferred examples. 
While these techniques effectively improve training stability, our empirical findings show that models trained using these regularization methods perform worse compared to those trained without such regularization.

\begin{figure}[t]
\centering
\begin{minipage}{.99\linewidth}
\centering
    \includegraphics[width = \linewidth]{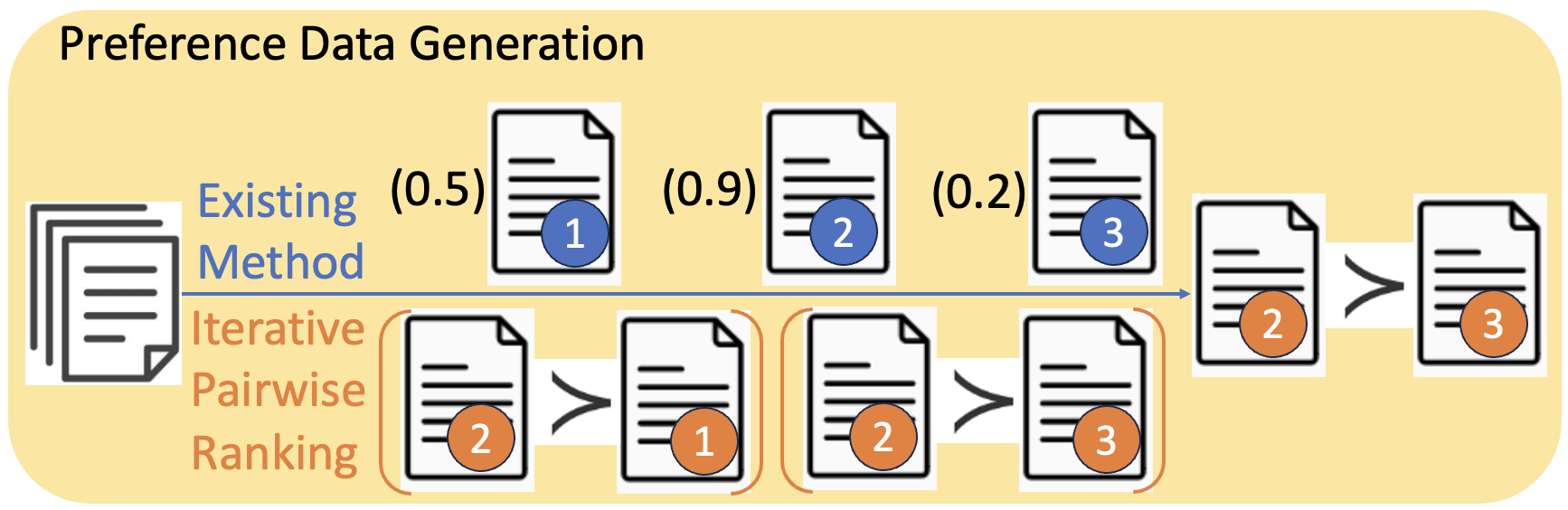}
\end{minipage}
\begin{minipage}{.99\linewidth}
\centering
    \includegraphics[width = \linewidth]{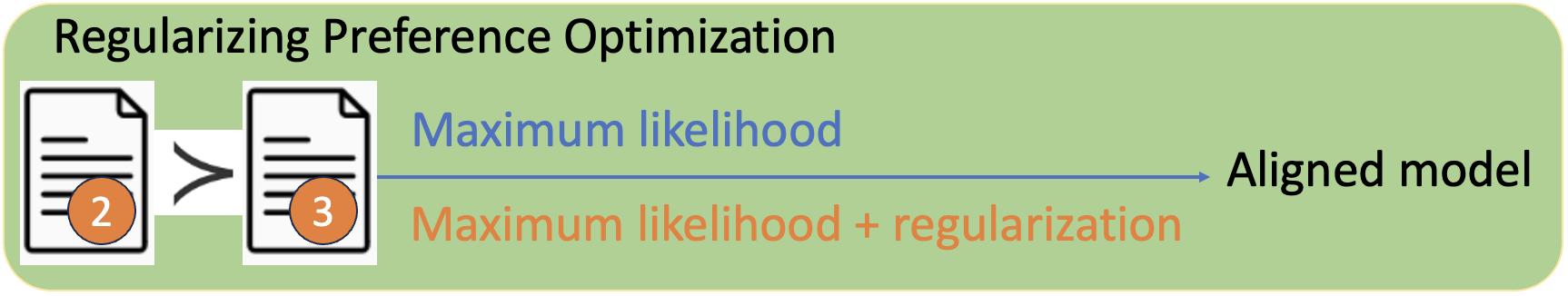}
\end{minipage}
\caption{
Overview for DPO pipeline.
Preference data generation: existing scoring-based methods select preferred and dispreferred completions based on a single score,
our proposed iterative pairwise ranking uses pairwise comparison signals to construct preference data.
Regularizing preference optimization: we propose a budget-controlled regularization that balances training stability and model alignment performance.
}
\label{fig: preference optimization overview}
\end{figure}

In this paper, we aim to improve the preference optimization pipeline. 
Our work introduces both high-quality preference data generation and improved regularization techniques to address the above limitations. 
Shown in 
Fig. \ref{fig: preference optimization overview}, 
we first propose an iterative pairwise ranking method to construct high-quality preference data.
Then we use this dataset to train a model with standard preference optimization objective augmented with a novel budget-controlled regularization.
The contributions of this work are  as follows:
\begin{itemize}
    \item 
    We introduce an iterative pairwise ranking mechanism that employs pairwise comparison signals to construct preference data.
    Specifically, given multiple completions for an input prompt, an LLM judge sequentially compares the previous winning completion with the next candidate until an optimal completion is found. 
    Empirical results demonstrate that preference data generated by our method consistently surpasses existing for both in-domain and out-of-distribution tasks.
    \item 
    We study the effects of supervised next-word prediction regularization and reveal that while this technique prevents significant reductions in target model's predicted likelihood of preferred examples,
    preference optimization tends to achieve better results when the likelihood of both preferred and dispreferred examples are slightly reduced.
    This observation leads to a novel budget-controlled regularization we propose,
    which controls the amount of reduction on target model's predicted likelihood of preferred completions.
    \item We demonstrate that integrating the above two designs yields preference aligned models that outperform the current SOTA across two widely-adopted benchmark evaluations.
\end{itemize}

\section{Preference Dataset Generation}
\label{sec: preference dataset generation}
The quality of preference data is crucial to the performance of any preference optimization algorithm.
This section first outlines existing preference data generation methods (Sec. \ref{sec: scoring-based ranking methods}), 
then introduces an iterative pairwise ranking approach (Sec. \ref{sec: iterative pair-wise ranking}).

\subsection{Existing Data Generation Methods}
\label{sec: scoring-based ranking methods}
A preference dataset consists of $N$ tuples $\{ (x^i, y^{i}_w, y^{i}_l) \}_{i=1}^N$,
where $x^i$, $y^{i}_w$ and $y^{i}_l$ represent input prompt, preferred and dispreferred completions, respectively.
In this work, we assume that input prompts are provided.
In an online setting, the target LLM generates multiple completions for each prompt, denoted as ${ y^{i, 1}, y^{i, 2},...,y^{i, M} }$.
Then preference data are constructed by selecting preferred and dispreferred completions from these candidates \citep{xiong2024iterative}.

Let $r^{\ast}(x, y)$ denote the ground-truth reward model that provides a reward score on a prompt-completion pair $(x, y)$.
The objective function for identifying the most preferred completion $y^i_w$ can be formulated as follows,
\begin{equation}
\label{eq: objective to select preferred completion}
\vspace{-2mm}
y^i_w 
= 
\arg \max\limits_{y \in \{ y^{i, m} \}_{m=1}^M } r^{\ast} (x^i, y).
\end{equation}
The same methodology can be applied to search for $y^i_l$ by considering the $\arg \min$ over $\{ y^{i, m} \}_{m=1}^M$.
Typically, Eq. \eqref{eq: objective to select preferred completion} is solved using an estimated reward model $r^{\phi}(x, y)$ \citep{pal2024smaug, feng2024towards, liu2024provably}. 
Then preferred and dispreferred completions are selected based on these estimated reward scores.
While these reward models demonstrate high accuracy on tasks closely aligned with their training datasets \citep{lambert2024rewardbench},
they generalize poorly on out-of-distribution tasks and require adaptation to new domains \citep{bai2022training, tang2024understanding}.

\subsection{Proposed: Iterative Pairwise Ranking via Dueling Bandits}
\label{sec: iterative pair-wise ranking}
We propose an \textbf{I}terative \textbf{P}airwise \textbf{R}anking (\textbf{IPR}) approach motivated by the dueling bandits framework \citep{sui2018advancements} to address Eq. \eqref{eq: objective to select preferred completion}.
This method searches for the preferred completion through sequential pairwise comparisons.

\paragraph{A simple dueling bandit algorithm for identifying preferred completion.}
~
Unlike the standard setting that requires absolute feedback for each candidate (e.g., using an estimated reward score as described in Sec. \ref{sec: scoring-based ranking methods}), the dueling bandits framework assumes the presence of only binary (or ternary if tie presents) feedback about the relative quality of each pair of candidates.

We begin by assuming the existence of a Condorcet winner \citep{urvoy2013generic}, which represents a unique optimal solution superior to all others. 
Typically, Copeland's method \citep{merlin1996copeland} is used to select the optimal candidate who wins the most pairwise comparisons, considering the possibility of ties. 
However, this method requires $\mathcal{O}(M^2)$ comparisons, making it computationally demanding. 
To improve efficiency, we introduce two assumptions to identify the winner:
\begin{enumerate}[noitemsep]
    \item
    \textbf{Transitive:}
    $y^{(i, a)} \succ y^{(i, b)}$ and $y^{(i, b)} \succ y^{(i, c)}$ leads to $y^{(i, a)} \succ y^{(i, c)}$ almost surely, where $a, b, c \in \{1, 2, \dots, M\}$.
    \item 
    \textbf{Symmetry:}
    The ordering of two completions does not affect the comparison result $W$,
    $W(x^i, y^{(i, a)}, y^{(i, b)}) = W(x^i, y^{(i, b)}, y^{(i, a)})$.
\end{enumerate}
Given these assumptions, identifying the most preferred completion from $M$ candidates can be accomplished from $(M-1)$ comparisons.
Specifically, the algorithm initiates by comparing the first pair of completions, followed by comparing their winner with the next candidate. 
This iterative process continues until an overall winner is determined.

\section{Regularizing Preference Optimization}
\label{sec: preference optimization regularizations}
In this section, we first analyze the failure mode associated with preference optimization algorithms (Sec. \ref{sec: failure mode of preference optimization}). 
We then discuss regularization techniques aimed at improving training stability (Sec. \ref{sec: regularization techniques}). 
Lastly, we introduce a budget-controlled regularization (Sec. \ref{sec: budget controlled regularization}) that balances between training stability and model alignment performance.

\subsection{Failure Mode of Preference Optimization}
\label{sec: failure mode of preference optimization}
Given a pairwise preference dataset,
DPO (and its variants) optimizes the LLM to increase the gap between the probabilities of generating preferred and dispreferred completions, subject to a KL-divergence constraint that prevents large deviation of the optimized model from the initial base model,
this is formulated as a maximum likelihood optimization of the target distribution $\pi_{\theta}(\cdot | x)$,
\begin{align}
\label{eq: dpo}
& \mathcal{L}_{DPO}(\pi_{\theta}, \pi_{\rm{ref}})
\nonumber \\
& =
- \mathbb{E}_{(x, y_w, y_l) \sim \mathcal{D}}
\big[
\log \sigma
\big(
r(x, y_w) - r(x, y_l)
\big)
\big],
\nonumber \\
&
\;\; {\rm where}
\;\;
r(x, y) = \beta \log \bigg( \frac{\pi_{\theta} (y | x)} {\pi_{\rm{ref}} (y | x)} \bigg),
\end{align}
where the reward function $r(x, y)$ is parameterized by the ratio between target and reference models scaled by a hyper-parameter $\beta$.
The DPO loss is a function of the difference in the log-ratios,
which means that we can achieve a low loss value even if the reward of preferred completion $r(x, y_w)$ is lowered,
as long as the reward of dispreferred completion $r(x, y_l)$ is sufficiently lower.
This implies that the log-likelihood of the preferred
completions can be reduced even below the original log-likelihood from the reference model.

\begin{figure*}[t]
\centering
\begin{minipage}{.68\columnwidth}
\centering
    \includegraphics[width = \linewidth]{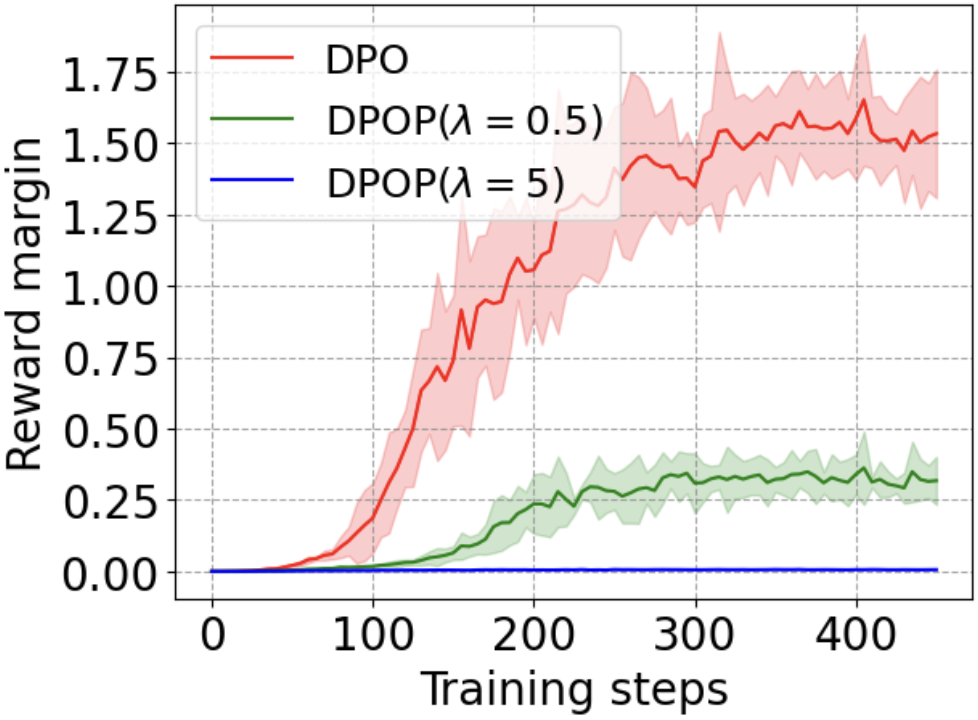}
    (a): $r(x, y_w) - r(x, y_l)$
\end{minipage}
\begin{minipage}{.68\columnwidth}
\centering
    \includegraphics[width = \linewidth]{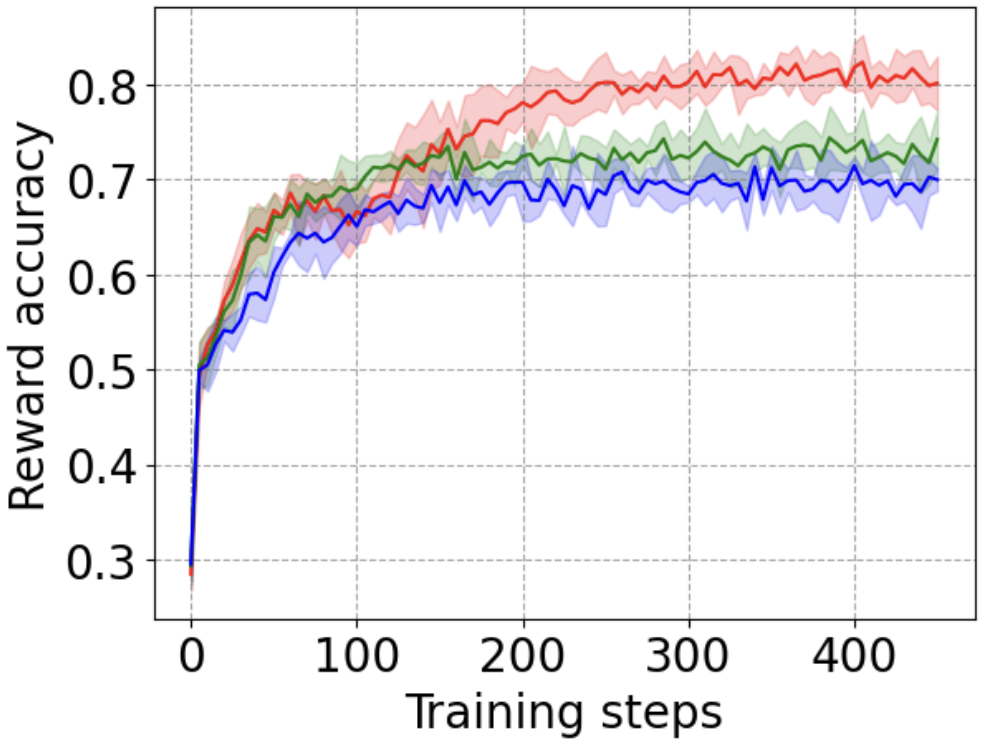}
    (b): Reward accuracy
\end{minipage}
\begin{minipage}{.68\columnwidth}
\centering
    \includegraphics[width = \linewidth]{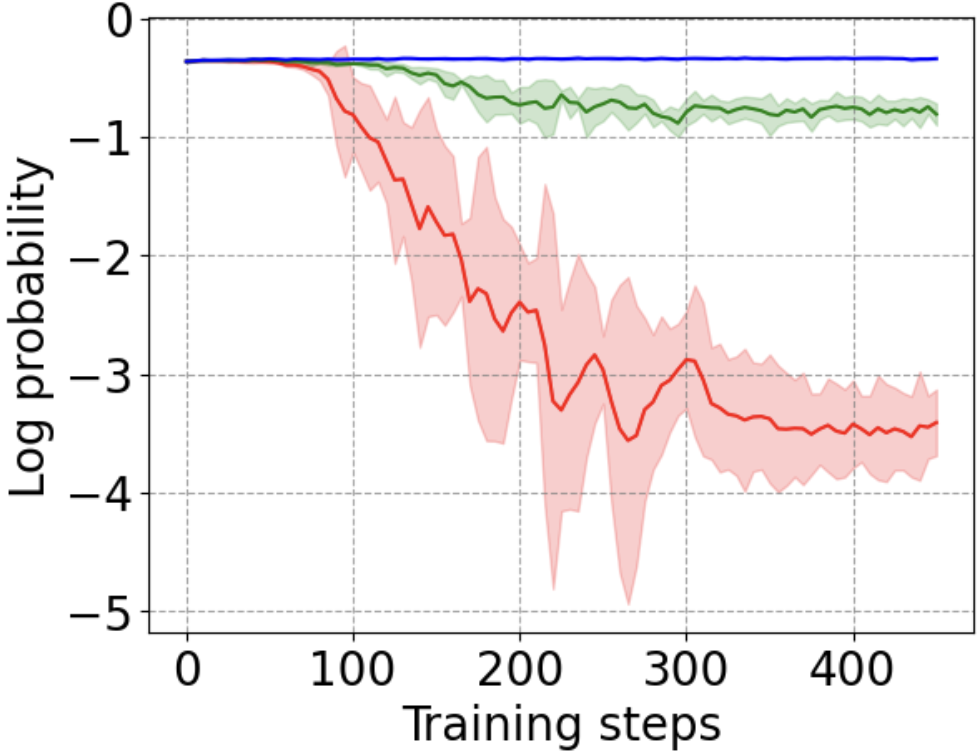}
    (c): $\log \pi_{\theta}(y_w | x)$
\end{minipage}
\caption{
Training progresses of DPO and DPOP.
(a) Reward margin: Measures the difference in rewards between preferred and dispreferred completions, which is the main objective in DPO training. 
(b) Reward accuracy: Shows the percentage of preferred completions that have higher rewards than their dispreferred ones. 
(c) Log probability: Indicates the average log-likelihood of preferred completions.
}
\label{fig: demo dpo reward margin+reward accuracy+log probability}
\end{figure*}

We empirically showcase the failure mode in preference optimization. 
Specifically, we apply DPO \citep{rafailov2024direct} to train the Llama-3.1-8B instruct model \href{https://huggingface.co/meta-llama/Llama-3.1-8B}{Llama-3.1-8B} using the UltraFeedback Binarized dataset \href{https://huggingface.co/datasets/openbmb/UltraFeedback}{UltraFeedback} (details in Sec. \ref{sec: experimental setup}). 
As shown in Fig. \ref{fig: demo dpo reward margin+reward accuracy+log probability}, while DPO effectively improves both the reward margin and reward accuracy, indicating that the model better learns the underlying preference data, there is a significant reduction in the log-likelihood of predicting preferred completions, leading to the failure mode.
Extensive numerical evidences on the failure mode of DPO (and its variants) across different settings can be found in Appendix \ref{appendix: preference optimization regularizations}.

\subsection{Next-Word Prediction Regularization}
\label{sec: regularization techniques}
Regularization for preference optimization has shown its effectiveness to prevent the failure mode.
These regularization techniques generally focus on a supervised next-word prediction objective with a goal of increasing the log-likelihood of predicting the preferred completions during training.
One notable algorithm is named DPO-Positive (DPOP) \citep{pal2024smaug},
\begin{align}
\label{eq: dpop}
& \mathcal{L}_{DPOP}(\pi_{\theta}, \pi_{\rm{ref}})
\nonumber \\
& = - \mathbb{E}_{(x, y_w, y_l) \sim \mathcal{D}}
\big[
\log \sigma
\big(
r(x, y_w) - r(x, y_l)
\nonumber \\
& \;\;\;\; -
\lambda \cdot 
\max
\big(0, \log \big(\frac{\pi_{\rm ref}(y_w|x)} {\pi_{\theta}(y_w|x)} \big)\big)
\big)
\big],
\end{align}
where $\lambda$ is a hyper-parameter to balance between the reward difference of DPO objective and regularization term.
The DPOP regularization can be interpreted as a reparameterization of the reward function for the preferred samples,
\begin{align*}
& r(x, y_w)
=
\beta \log \bigg( \frac{\pi_{\theta} (y | x)} {\pi_{\rm{ref}} (y | x)} \bigg)
\nonumber \\
& \;\;\;\; -
\lambda \cdot
\max
\big(0, \log \big(\frac{\pi_{\rm ref}(y_w|x)} {\pi_{\theta}(y_w|x)} \big)\big),
\end{align*}
then it optimizes the pairwise preferences, $r(x, y_w)-r(x, y_l)$, via a Bradley-Terry (BT) model \citep{david1963method}. 
The results of DPOP is illustrated in Fig. \ref{fig: demo dpo reward margin+reward accuracy+log probability}.
As can be seen, with a sufficiently large $\lambda$ (e.g., $\lambda=5$),
DPOP addresses the failure mode of DPO by ensuring that the log-likelihood of preferred completions remains non-decreasing throughout the whole training process.

However, the DPOP approach of applying regularization inside the log-sigmoid function can be problematic with deterministic or near-deterministic preference data (e.g., the probability of $y_w \succ y_l$ is near $1$). 
This method tends to overfit the preference dataset, neglecting the KL-regularization term \citep{azar2024general}, which ultimately reduces the probability of accurately predicting the preferred completion.

\subsection{Budget Controlled Regularization}
\label{sec: budget controlled regularization}
Here we propose a \textbf{B}udget \textbf{C}ontrolled \textbf{R}egularization (\textbf{\methodA}) that balances between training stability and model alignment performance. 
First, similar to Contrastive Preference Optimization \cite{xucontrastive}, the proposed regularization acts as the supervised next-word prediction objective outside of the log-sigmoid function, which prevents the failure mode of DPO more effectively than DPOP by avoiding the overfitting issue.
Moreover, the analyses in Fig. \ref{fig: demo dpo reward margin+reward accuracy+log probability} reveal that the reduction in the log-likelihood of predicting preferred completions is necessary for the model to achieve a high reward margin and accuracy.
Specifically, as the regularization effect of DPOP strengthens (with an increase in $\lambda$), the resulting models underperform compared to those trained with DPO.
Extensive empirical validations can be found in Sec. \ref{appendix: preference optimization regularizations}.

\begin{figure}[t]
\centering
\begin{minipage}{.9\columnwidth}
\centering
    \includegraphics[width = \linewidth]{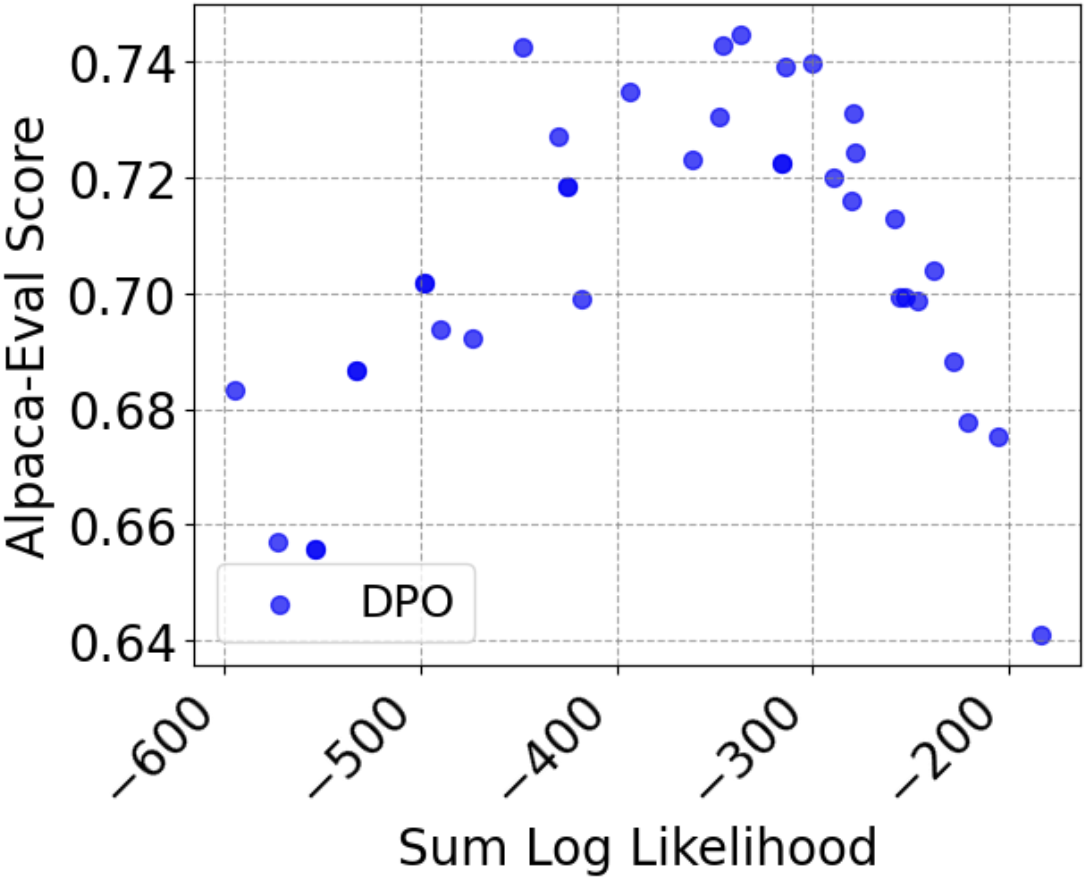}
\end{minipage}
\caption{
Optimization budget (log-likelihood of preferred completions) versus Alpaca-Eval win rate score.
Each point corresponds to a model trained on a particular set of hyperparameters.
}
\label{fig: llama preference optimization alpaca-eval win rate}
\end{figure}

Fig. \ref{fig: llama preference optimization alpaca-eval win rate} illustrates the trade-off between the average sum log-likelihood of preferred completions and model performance on the Alpaca-Eval $2.0$ dataset \citep{dubois2024length}. 
Each data point represents the evaluation result of a model checkpoint trained on a particular set of hyperparameters. The sum log-likelihood is averaged across the samples in dev set, while model performance is measured as the win rate against a golden reference completion. As training progresses, the sum log-likelihood decreases, consistent with Fig \ref{fig: demo dpo reward margin+reward accuracy+log probability}(c). The model performance initially improves but later declines due to overfitting to the preference dataset.
Thus, the regularization term should allow a certain reduction of the log-likelihood on preferred completion (defined as budget) for the decrease in sum log-likelihood but penalize the decrease beyond the budget. The training objective with the proposed budget controlled regularization is as follows:
\begin{align}
\label{eq: relaxed regularized dpo}
& \mathcal{L}_{DPO\methodA}(\pi_{\theta}, \pi_{\rm{ref}}) = \mathcal{L}_{DPO}(\pi_{\theta}, \pi_{\rm{ref}})
\nonumber \\
&
+
\lambda \mathbb{E}_{(x, y_w) \sim \mathcal{D}}
\max
\bigg(0, \log \frac{\pi_{\rm{ref}}(y_w | x)} {\pi_{\theta} (y_w | x)} - \delta
\bigg)
\end{align}
where $\delta$ is an non-negative hyper-parameter.
Specifically,
when $\delta=0$, DPO{\methodA} strictly penalizes any reduction of likelihood of predicting the preferred completion.
A small positive $\delta$ allows the probability of predicting preferred completions to be slightly reduced, 
while maximizing the reward margin via $\mathcal{L}_{DPO}$.
Such regularization term enables the optimization process to achieve best trade-offs between the sum log-likelihood and policy performance.

\section{Related Works}
\label{sec: related works}
In this section, we outline preference optimization algorithms and existing regularization techniques to improve training stability.
Extensive discussion is provided in Appendix \ref{sec: appendix related works}.

\paragraph{DPO and Its Variants.}
~Since the introduction of DPO \citep{rafailov2024direct}, several algorithms have emerged to further refine preference optimization. 
SimPO (Simple Preference Optimization) introduces length regularization on the log-probabilities of both preferred and dispreferred completions, eliminating the need for a reference model \citep{meng2024simpo}. 
IPO (Identity Preference Optimization) addresses the shortcomings of BT preference modeling in cases where preference data are highly deterministic, when the preferred completion is almost always better to the dispreferred one. 
In such cases, the KL-divergence regularization becomes ineffective.
IPO resolves this by replacing the logistic loss with a squared loss and incorporating a margin, providing a more theoretically sound approach \citep{azar2024general}.
Other notable algorithms include 
RPO (Regularized preference optimization) that emphasizes the role of length regularization \citep{park2024disentangling},
and iterative preference learning that iteratively refine the target LLM based on preference data \citep{xiong2024iterative, kim2024sdpo}.

\paragraph{Supervised Next-Word Prediction Regularization Improves Training Stability.}
~To improve the training stability of preference optimization, various forms of supervised next-word prediction regularization have been proposed to improve training stability. 
SLIC (sequence likelihood calibration) adds a term to maximize log-likelihoods on certain reference completions \citep{zhao2023slic}, CPO (Contrastive Preference Optimization) applies a behavior cloning regularizer \citep{hejnacontrastive,xucontrastive}.
Additionally, DPOP introduces a hinge loss on the log-ratio between the reference and target models \citep{pal2024smaug}. 
Despite the improvements in training stability, our analysis indicates that regularized preference optimization often results in worse performance compared to non-regularized approaches.

\section{Experimental Results}
\label{sec: experimental results}
In this section, we showcase the improved model alignment performance achieved through the proposed designs (Sec. \ref{sec: main results summary}). 
Additionally, we provide a comprehensive ablation study to assess the quality of preference data generated by \methodB~ and the effectiveness of \methodA (Sec. \ref{sec: ablation study}).

\subsection{Experimental Setup}
\label{sec: experimental setup}
We discuss our design choices regarding base models, training details and evaluation metrics.
Additional details are provided in Appendix \ref{sec: additional experimental setup}.

\paragraph{Base models.}
~We conduct all experiments using both
\href{https://huggingface.co/meta-llama/Llama-3.1-8B}{Llama-3.1-8B instruct} and
\href{https://huggingface.co/mistralai/Mistral-7B-Instruct-v0.2}{Mistral-Instruct-7B}.
Both models have undergone extensive instruction-tuning. 

\paragraph{Preference Data Construction.}
~To mitigate the distribution shift between base models and the preference optimization process, we generate the preference dataset using the base models \citep{tang2024understanding, meng2024simpo, xiong2024iterative}.
This makes the training process closer to an on-policy setting.
Specifically, we use prompts from the UltraFeedback dataset \citep{cui2023ultrafeedback} and regenerate the preferred and dispreferred completions with the base models.
For each prompt, as a default setting, we generate $5$ completions using the base model with a sampling temperature of $0.8$.
For reward model-based method, we consider \href{https://huggingface.co/RLHFlow/ArmoRM-Llama3-8B-v0.1}{ArmoRM-Llama3-8B-v0.1} \citep{ArmoRM} to score all completions and select the highest-scoring one as $y_w$ and the lowest-scoring one as $y_l$.
In addition, we construct another high-quality preference dataset using the proposed \methodB.

\paragraph{Training details.}
~We apply full-parameter training and search for the optimal learning rate from $1e^{-6}$ to $8e^{-6}$.
All training runs apply a fixed batch size of $128$ and max epoch of $1$.

\begin{table*}[h!]
\centering
\begin{tabular}{l|cc}
\thickhline
\multicolumn{1}{c|}{Method}  & Objective Function \\ \hline
DPO & 
$- \log \sigma \Big(\beta \log \big(\frac{\pi_{\theta}(y_w|x)} {\pi_{\rm ref}(y_w|x)} \big) -\beta \log \big(\frac{\pi_{\theta}(y_l|x)} {\pi_{\rm ref}(y_l|x)} \big) \Big)$
\\
IPO &
$- \Big(\log \big(\frac{\pi_{\theta}(y_w|x)} {\pi_{\rm ref}(y_w|x)} \big) - \log \big(\frac{\pi_{\theta}(y_l|x)} {\pi_{\rm ref}(y_l|x)} \big) - \frac{1} {2 \tau} \Big)^2$\\

SimPO &
$- \log \sigma \Big(\frac{\beta} {|y_w|} \log \pi_{\theta} (y_w| x) -  \frac{\beta} {|y_l|} \log \pi_{\theta} (y_l| x) - \gamma \Big)$ \\
 \thickhline
DPO\textcolor{blue}{\methodA} & 
$- \log \sigma \Big(\beta \log \big(\frac{\pi_{\theta}(y_w|x)} {\pi_{\rm ref}(y_w|x)} \big) -\beta \log \big(\frac{\pi_{\theta}(y_l|x)} {\pi_{\rm ref}(y_l|x)} \big) \Big) 
\textcolor{blue}
{+
\lambda \cdot
\max
\bigg(0, \log \frac{\pi_{\rm{ref}}(y_w | x)} {\pi_{\theta} (y_w | x)} - \delta
\bigg)
}$ \\
IPO\textcolor{blue}{\methodA} & 
$- \Big(\log \big(\frac{\pi_{\theta}(y_w|x)} {\pi_{\rm ref}(y_w|x)} \big) - \log \big(\frac{\pi_{\theta}(y_l|x)} {\pi_{\rm ref}(y_l|x)} \big) - \frac{1} {2 \tau} \Big)^2
\textcolor{blue}
{+
\lambda \cdot
\max
\bigg(0, \log \frac{\pi_{\rm{ref}}(y_w | x)} {\pi_{\theta} (y_w | x)} - \delta
\bigg)
}$ \\
SimPO\textcolor{blue}{\methodA} & 
$- \log \sigma \Big(\frac{\beta} {|y_w|} \log \pi_{\theta} (y_w| x) -  \frac{\beta} {|y_l|} \log \pi_{\theta} (y_l| x) - \gamma \Big) 
\textcolor{blue}
{+
\lambda \cdot
\max
\bigg(0, -\frac{\log \pi_{\theta} (y_w | x)}{|y_w|} - \delta
\bigg)
}$ \\
\thickhline
CPO &
$- \log \sigma \Big(\frac{\beta} {|y_w|} \log \pi_{\theta} (y_w| x) -  \frac{\beta} {|y_l|} \log \pi_{\theta} (y_l| x) - \gamma \Big) - \frac{\lambda} {|y_w|}  \log \pi_{\theta} (y_w | x)$ \\
DPOP &
$- \log \sigma \Big(\beta \log \big(\frac{\pi_{\theta}(y_w|x)} {\pi_{\rm ref}(y_w|x)} \big) -\beta \log \big(\frac{\pi_{\theta}(y_l|x)} {\pi_{\rm ref}(y_l|x)}  \big) - \lambda \cdot \max \big(0, \log \big(\frac{\pi_{\rm ref}(y_w|x)} {\pi_{\theta}(y_w|x)} \big)\big) \Big)$ \\
 \thickhline
\end{tabular}
\caption{
Preference optimization algorithms and their objective function implementations.
}
\label{table: preference optimization baseline algorithms}
\end{table*}

We summarize all baseline algorithms in Table \ref{table: preference optimization baseline algorithms}. 
As baseline algorithms, we implement \textbf{DPO},
\textbf{IPO},
\textbf{SimPO},
\textbf{CPO} and \textbf{DPOP}.
In addition, we apply the proposed \methodA~ to DPO, IPO, and SimPO,
which lead to \textbf{DPO\methodA}, \textbf{IPO\methodA}, and \textbf{SimPO\methodA}, respectively.
Notice that SimPO\methodA~ retains the advantage of SimPO by not requiring a reference model during training, and its budget-controlled regularization focuses solely on the log likelihood of preferred completions from the target model.

\paragraph{Evaluations.}
~All winrate-based evaluations are done using \href{https://huggingface.co/prometheus-eval/prometheus-bgb-8x7b-v2.0}{Mixtral-8x7B-Instruct } as the model judge \citep{kim2024prometheus}.
To evaluate the performance of aligned models,
we use two popular open-ended instruction-following benchmarks:
AlpacaEval $2.0$ \citep{dubois2024length} and Arena-Hard \citep{li2024live}.
These benchmarks assess the model's versatile conversational capabilities across a wide range of queries and have been widely adopted by the community.

In addition,
all experiments are done using $8$ A100 GPUs.

\subsection{Main Results Summary}
\label{sec: main results summary}
\begin{table*}[h!]
\centering
\begin{tabular}{l|cccccccc}
\thickhline
\thickhline
\multicolumn{9}{c}{\textbf{Llama-3.1-Instruct (8B)}} \\
\thickhline
\multicolumn{9}{c}{Alpaca-Eval $2.0$ (Base Model: 47.64)} \\ \hline
 & DPO & IPO & SimPO & DPO\textcolor{blue}{\methodA} & IPO\textcolor{blue}{\methodA} & SimPO\textcolor{blue}{\methodA} & CPO & DPOP \\ 
Armo Llama3 & 58.07 & 57.00 & 65.16 & / & / & / & 55.71 & 48.94 \\ 
\methodB(Llama8B) & 72.86 & 69.94 & 66.77 & / & / & / & \textbf{82.86} & \textbf{54.66} \\
\methodB(Llama70B) & \textbf{73.11} & \textbf{71.30} & \textbf{85.32} & \textbf{74.35} & \textbf{72.92} & \textbf{85.90} & 79.69 & 54.16 \\ \hline

\multicolumn{9}{c}{Arena-Hard (Base Model: 71.44)} \\ \hline
 & DPO & IPO & SimPO & DPO\textcolor{blue}{\methodA} & IPO\textcolor{blue}{\methodA} & SimPO\textcolor{blue}{\methodA} & CPO & DPOP \\
Armo Llama3 & 79.90 & 78.10 & 84.10 & / & / & / & 74.00 & 71.30 \\
\methodB(Llama8B) & \textbf{80.70} & \textbf{82.40} & 80.00 & / & / & / & \textbf{85.90} & 71.60 \\
\methodB(Llama70B) & 80.50 & 80.40 & \textbf{89.30} & 79.30 & 79.50 & \textbf{89.30} & 83.37 & \textbf{73.90} \\

\thickhline
\multicolumn{9}{c}{\textbf{Mistral-Instruct (7B)}} \\
\thickhline
\multicolumn{9}{c}{Alpaca-Eval $2.0$ (Base Model: 25.03)} \\ \hline
 & DPO & IPO & SimPO & DPO\textcolor{blue}{\methodA} & IPO\textcolor{blue}{\methodA} & SimPO\textcolor{blue}{\methodA} & CPO & DPOP \\
Armo Llama3 & $38.14^{\ast}$ & $36.27^{\ast}$ & $49.94^{\ast}$ & / & / & / & 28.79 & 28.70 \\
\methodB(Llama8B) & 60.34 & 58.30 & 57.35 & / & / & / & 47.39 & \textbf{41.98} \\
\methodB(Llama70B) & \textbf{67.75} & \textbf{65.49} & \textbf{61.06} & 67.40 & \textbf{65.52} & \textbf{64.99} & \textbf{48.63} & 41.28 \\ \hline

\multicolumn{9}{c}{Arena-Hard (Base Model: 56.70)} \\ \hline
 & DPO & IPO & SimPO & DPO\textcolor{blue}{\methodA} & IPO\textcolor{blue}{\methodA} & SimPO\textcolor{blue}{\methodA} & CPO & DPOP \\
Armo Llama3 & $67.13^{\ast}$ & $61.60^{\ast}$ & $\textbf{72.04}^{\ast}$ & / & / & / & 62.00 & 62.90 \\
\methodB(Llama8B) & 68.70 & 65.20 & 67.40 & / & / & / & 67.20 & \textbf{66.93} \\
\methodB(Llama70B) & \textbf{71.80} & \textbf{71.70} & 70.84 & 71.53 & 71.20 & 63.10 & \textbf{71.10} & 65.40 \\
\thickhline
\end{tabular}
\caption{
AlpacaEval 2 \citep{dubois2024length} and Arena-Hard \citep{li2024live} evaluation results for preference-tuned \textbf{Llama-3.1 (8B)} and \textbf{Mistral-Instruct (7B)} models. 
\textbf{Armo Llama3} applies
\href{https://huggingface.co/RLHFlow/ArmoRM-Llama3-8B-v0.1}{ArmoRM-Llama3-8B-v0.1} to construct preference data,
\methodB(Llama8B) and \methodB(Llama70B) apply the proposed iterative pairwise ranking with Llama-3.1 8B and Llama-3.1 70B, respectively.
$x^{\ast}$ indicates that the scores are obtained from public models.
\methodA~ is only applied to train on the highest-quality preference data generated from \methodB(Llama70B).
}
\label{table: model performance evaluation llama and mistral}
\end{table*}

Table \ref{table: model performance evaluation llama and mistral} summarizes the alignment performance of all trained models.
\paragraph{Preference optimization with \methodB~ significantly outperforms existing methods.}
~By comparing models trained using the reward model (Armo Llama3),
using \methodB~ method to construct preference data significantly improves model alignment performance across different preference optimization algorithms.
In the Alpaca-Eval $2.0$ evaluation, the Llama-3.1 models trained with DPO and SimPO show substantial performance gains, with winrate improvements of $15 \%$ and $20 \%$, respectively, when trained with \methodB~ preference data. 
Notably, models trained with regularized objectives like CPO exhibit an even greater winrate increase of $27 \%$.
This performance improvement can be seen for preference tuned Mistral-Instruct (7B).
Furthermore, the effectiveness of the \methodB~ method is influenced by the capability of the LLM used as the preference judge. 
Models trained with \methodB~ data constructed from the Llama70B (denoted as \methodB(Llama70B)) outperform those using the Llama8B judge (denoted as \methodB(Llama8B)), underscoring the importance of the judge model's quality in constructing high-performing preference datasets.

\paragraph{\methodA~ matches state-of-the-art performance with less optimization budget.}

\begin{figure}[t]
\centering
\begin{minipage}{.9\columnwidth}
\centering
    \includegraphics[width = \linewidth]{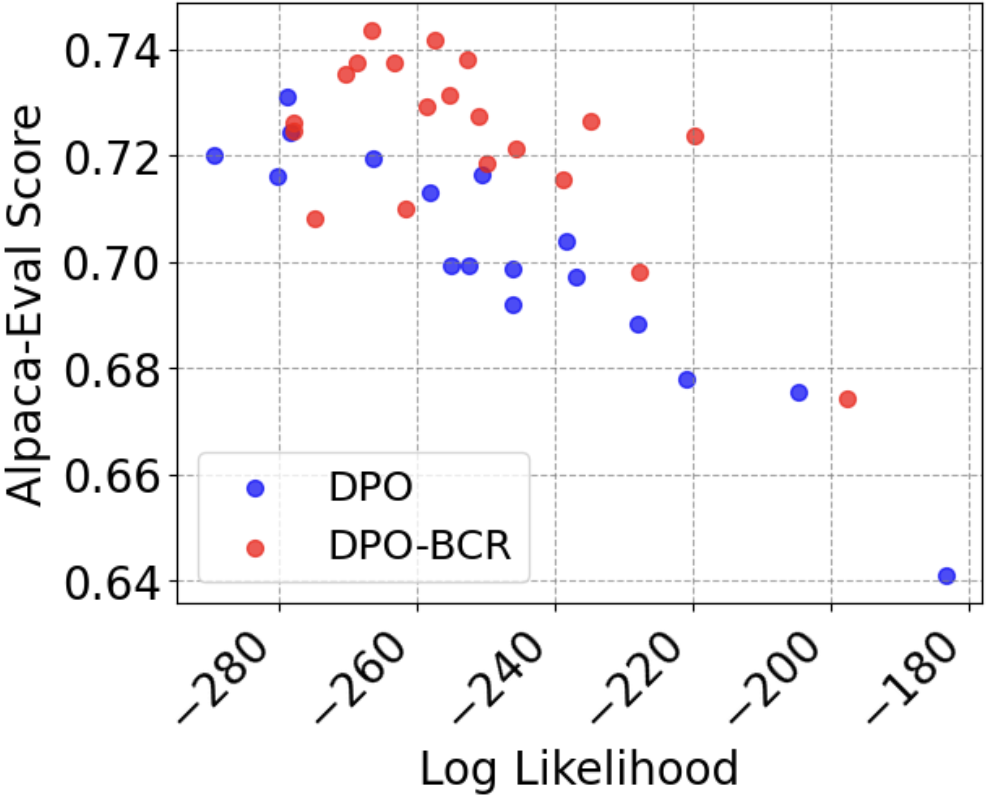}
    (a): Llama-3.1(8B) DPO
\end{minipage}
\begin{minipage}{.9\columnwidth}
\centering
    \includegraphics[width = \linewidth]{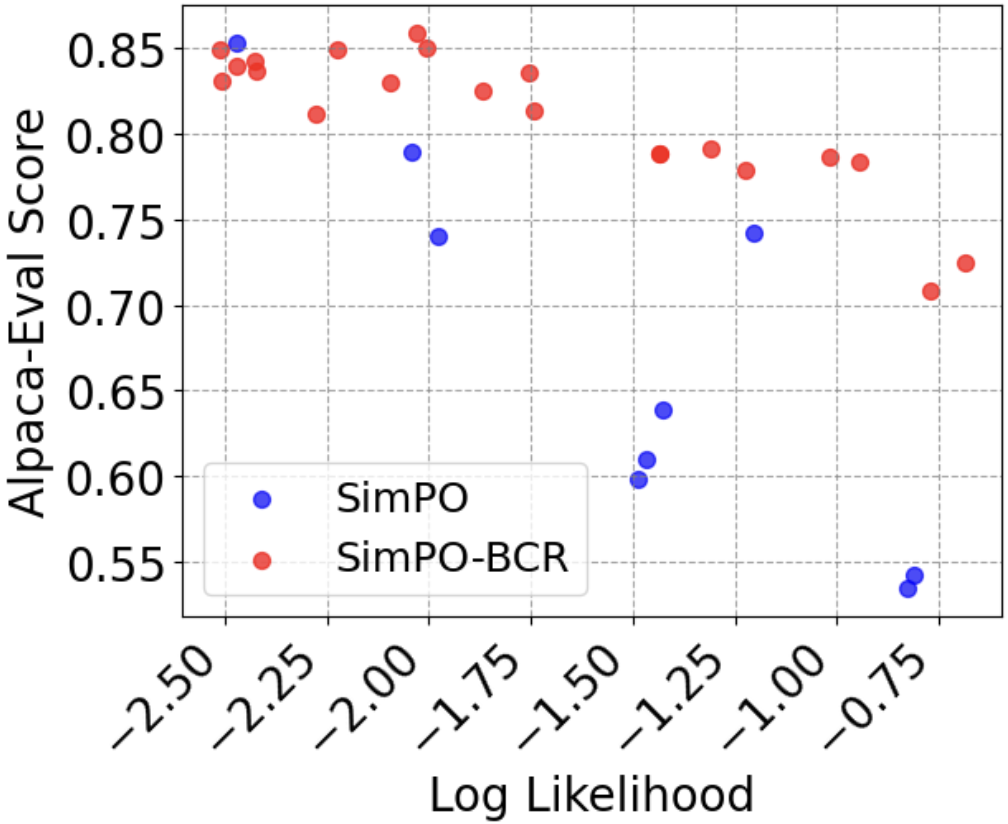}
    (b): Llama-3.1(8B) SimPO
\end{minipage}
\caption{
Optimization budget (log-likelihood of preferred completions) versus Alpaca-Eval.
(a) DPO versus DPO-\methodA: sum of log-likelihood of preferred completions is used.
(b) SimPO versus SimPO-\methodA: average of log-likelihood of preferred completions is used.
}
\label{fig: llama preference optimization vs bcr}
\end{figure}

~Recall in Sec. \ref{sec: failure mode of preference optimization}, as both reward margin and reward accuracy increase, the log-likelihood of predicting preferred completions decreases, indicating the failure mode of preference optimization. 
Here we define the optimization budget as the log-likelihood of predicting preferred samples. 
As shown, with models trained using \methodB, while adding \methodA~ for preference optimization does not significantly further improve model alignment performance, it allows the trained model to achieve the same level of performance using much less optimization budget. 
Specifically, 
for Llama-3.1-Instruct (8B), SimPO\methodA~ outperforms SimPO by increasing the score from $85.3\%$ to $85.9\%$, 
as shown in Fig. \ref{fig: llama preference optimization vs bcr} (b),
SimPO\methodA~ reduces the optimization budget to $2.03$ compared to the $2.47$ required by naive SimPO.

\subsection{Ablation Study}
\label{sec: ablation study}
\begin{table}[h!]
\centering
\begin{tabular}{l|ccc}
\thickhline
\multicolumn{3}{c}{Ultrafeedback Preference Data Quality} \\ \hline
 & Llama-3.1 & Mistral \\ \hline
Reward Gemma & 76.50 & 71.77 \\
Armo Llama-3 & 75.31 & 68.57 \\
Urm Llama-3.1 & 67.60 & 57.86 \\
Llama-3.1 (70B) & 66.40 & 73.33 \\ \thickhline
\methodB(Llama-3.1 8B) & 79.50 & 82.45 \\
\methodB(Llama-3.1 70B) & \textcolor{blue}{82.33} & \textcolor{blue}{86.53} \\
\thickhline
\end{tabular}
\caption{
The scores represent the agreement (in $\%$) with the model judge (\href{https://huggingface.co/prometheus-eval/prometheus-bgb-8x7b-v2.0}{Mixtral-8x7B-Instruct}) by using the dispreferred completion as the baseline and the preferred completion as the candidate.
Scores in columns $1$ and $2$ use completions generated from \textbf{Llama-3.1 (8B)} and \textbf{Mistral-Instruct (7B)}, respectively.
}
\label{table: ultrafeedback preference data quality evaluation}
\end{table}

\begin{table}[h!]
\centering
\begin{tabular}{l|cccccc}
\thickhline
\multicolumn{3}{c}{Out-Distribution Preference Data Quality} \\ \hline
 & MsMarco & PubMed \\ \hline
Reward Gemma & 70.32 & 68.86 \\
Armo Llama-3 & 57.81 & 58.85 \\
Urm Llama-3.1 & 39.60 & 44.81 \\
Llama-3.1 (70B) & 70.51 & 70.59 \\ 
\thickhline 
\methodB(Llama-3.1 70B) & \textcolor{blue}{81.61} & \textcolor{blue}{83.01} \\
\thickhline
\end{tabular}
\caption{
The scores represent the agreement (in $\%$) with the model judge (\href{https://huggingface.co/prometheus-eval/prometheus-bgb-8x7b-v2.0}{Mixtral-8x7B-Instruct}) by using the dispreferred completion as the baseline and the preferred completion as the candidate.
Completions are generated using \textbf{Llama-3.1 (8B)}.
}
\label{table: out-distribution preference data quality evaluation}
\end{table}

\paragraph{\methodB~ results in high quality preference data.}
~We perform a preference data quality analysis using three public reward models listed at top of the RewardBench \citep{lambert2024rewardbench}: \href{https://huggingface.co/Skywork/Skywork-Reward-Gemma-2-27B}{Reward Gemma} \citep{skyworkreward2024}, \href{https://huggingface.co/RLHFlow/ArmoRM-Llama3-8B-v0.1}{Armo Llama-3} \citep{ArmoRM}, and \href{https://huggingface.co/LxzGordon/URM-LLaMa-3.1-8B}{Urm Llama-3.1} \citep{wang2024arithmetic}. 
Additionally, we use Llama-3.1 (70B) to select preferred and dispreferred completions from all candidates.
Compared to \methodB,
this generation-based approach directly selects the most preferred completion from all candidate completions, without using sequential pairwise comparison signals.

Table \ref{table: ultrafeedback preference data quality evaluation} summarizes the analysis of preference data quality on Ultrafeedback. 
When using Llama-3.1 as the base model to generate completions, \methodB(Llama-3.1 70B) achieves an agreement score of $82.33 \%$ with the model judge, while the reward model, such as Armo Llama-3, only reaches $75.31 \%$ agreement. 
This validates the performance improvement in Table \ref{table: model performance evaluation llama and mistral}, comparing models trained using Armo Llama-3 and \methodB(Llama-3.1 70B).

For out-of-distribution tasks,
Table \ref{table: out-distribution preference data quality evaluation} summarizes the analysis of preference data quality on 
\href{https://huggingface.co/datasets/rungalileo/ragbench}{MsMarco} and \href{https://huggingface.co/datasets/rungalileo/ragbench}{PubMedQA}.
Specifically, on MsMarco, reward models achieve around $50 \%$ agreement, which is equivalent to random selection. 
The direct generation method suffers from positional bias, often favoring the first candidate, resulting in $70.5 \%$ agreement with the model judge. 
In contrast, \methodB~ produces high-quality preference data, with agreements of $81.6 \%$ on MsMarco and $83.01 \%$ on PubMedQA.

\paragraph{\methodA~ produces high-performing models with low optimization budget.}
~In Fig. \ref{fig: llama preference optimization vs bcr}, we show that the proposed \methodA~ results in high-performing models with low budget (smaller reduction on the log-likelihood of preferred completions).
For both vanilla DPO(SimPO) and proposed DPO\methodA(SimPO\methodA) algorithms,
the x-axis represents the average sum log-likelihood of preferred completions for DPO, and the average log likelihood normalized by completion length for SimPO. The y-axis shows model performance, defined as the win-rate against a golden reference completion on the Alpaca-Eval.
Each data point represents a model trained with specific hyperparameters. 
As can be seen,
at low-budget regime (larger log-likelihood), the proposed \methodA~ leads to significantly improved model performance.
In addition,
the regularization term significantly improves stability across different hyperparameters and outperforms vanilla versions at the same low budget regime. This is because the budget controlled regularization prevents overfitting to preference datasets and encourages finding the best solution within the allocated log-likelihood budget.

\section{Conclusion}
This work presents a comprehensive study of preference optimization algorithms, with a focus on improving preference data generation and regularization techniques.
Our empirical results show that preference optimization can be more effective when the likelihood of both preferred and dispreferred completions is managed carefully, allowing for a more balanced optimization. 
By combining \methodB~ for data generation and \methodA~ for preference optimization,
we demonstrate notable improvements.
There has been evidence that online alignment algorithms generally outperform offline methods,
we aim to extend the current pipeline to an online setting where the completions are generated during training by the target model.
We believe that the proposed designs can benefit the online setting with higher preference data quality and training stability.

\newpage

\section*{Ethical Considerations}
While \methodA~ and \methodB~ build up an effective preference optimization workflow, aligning LLM with human preferences raises certain ethical concerns.  One concern is that human preferences are complex, nuanced, and often contradictory. Attempting to codify human values into an AI system may oversimplify complex issues, for instance, it is difficult to decide whose preferences should be optimized for - the developers', users', or society's as a whole. Optimizing for any one group's preferences could lead to issues like bias and exclusion of minority viewpoints.

\section*{Limitations}
The proposed \methodB~ strategy for constructing preference data requires substantial computing resources. 
This is because it involves running multiple iterations of inferences with a large-scale LLM to select the preferred completion, and this process is repeated for all training data.



\bibliography{acl_latex}

\appendix


\section{Experimental Setup}
\label{sec: additional experimental setup}

\subsection{Training Details}
\label{sec: training details}

\paragraph{Training hyperparameters:}
Our findings highlight the critical role of hyperparameter tuning in achieving optimal performance for preference optimization methods. 
However, prior research may have underestimated its significance, potentially resulting in suboptimal baseline results. 
To ensure a fair comparison, we perform comprehensive hyperparameter tuning for all methods evaluated in our experiments.
Table \ref{table: preference optimization baseline algorithms implementation details} summarizes all hyperparameters used for all preference optimization algorithms.

\begin{table*}[h!]
\centering
\begin{tabular}{l|cc}
\thickhline
\multicolumn{1}{c|}{Method}  & Objective Function \\ \hline
\multirow{2}{*}{DPO} & 
$- \log \sigma \Big(\beta \log \big(\frac{\pi_{\theta}(y_w|x)} {\pi_{\rm ref}(y_w|x)} \big) -\beta \log \big(\frac{\pi_{\theta}(y_l|x)} {\pi_{\rm ref}(y_l|x)} \big) \Big)$
\\
& $\beta \in [0.01, 0.1]$
\\ \hline
\multirow{2}{*}{IPO} &
$- \Big(\log \big(\frac{\pi_{\theta}(y_w|x)} {\pi_{\rm ref}(y_w|x)} \big) - \log \big(\frac{\pi_{\theta}(y_l|x)} {\pi_{\rm ref}(y_l|x)} \big) - \frac{1} {2 \tau} \Big)^2$
\\
& $\tau \in [0.01, 0.1, 1]$
\\ \hline
\multirow{2}{*}{SimPO} &
$- \log \sigma \Big(\frac{\beta} {|y_w|} \log \pi_{\theta} (y_w| x) -  \frac{\beta} {|y_l|} \log \pi_{\theta} (y_l| x) - \gamma \Big)$ \\
& $\beta \in [2.5, 5, 10]$, $\gamma \in [0.1, 0.5]$
\\
 \thickhline
\multirow{2}{*}{DPO\textcolor{blue}{\methodA}} & 
$- \log \sigma \Big(\beta \log \big(\frac{\pi_{\theta}(y_w|x)} {\pi_{\rm ref}(y_w|x)} \big) -\beta \log \big(\frac{\pi_{\theta}(y_l|x)} {\pi_{\rm ref}(y_l|x)} \big) \Big) 
\textcolor{blue}
{+
\lambda \cdot
\max
\bigg(0, \log \frac{\pi_{\rm{ref}}(y_w | x)} {\pi_{\theta} (y_w | x)} - \delta
\bigg)
}$ \\
& $\beta \in [0.01, 0.1]$, $\lambda=1$, $\delta \in [1, 2, 4, 6, 8]$
\\ \hline
\multirow{2}{*}{IPO\textcolor{blue}{\methodA}} & 
$- \Big(\log \big(\frac{\pi_{\theta}(y_w|x)} {\pi_{\rm ref}(y_w|x)} \big) - \log \big(\frac{\pi_{\theta}(y_l|x)} {\pi_{\rm ref}(y_l|x)} \big) - \frac{1} {2 \tau} \Big)^2
\textcolor{blue}
{+
\lambda \cdot
\max
\bigg(0, \log \frac{\pi_{\rm{ref}}(y_w | x)} {\pi_{\theta} (y_w | x)} - \delta
\bigg)
}$ \\
& $\tau \in [0.01, 0.1, 1]$, $\delta \in [1, 2, 4, 6, 8]$
\\ \hline
\multirow{2}{*}{SimPO\textcolor{blue}{\methodA}} & 
$- \log \sigma \Big(\frac{\beta} {|y_w|} \log \pi_{\theta} (y_w| x) -  \frac{\beta} {|y_l|} \log \pi_{\theta} (y_l| x) - \gamma \Big) 
\textcolor{blue}
{+
\lambda \cdot
\max
\bigg(0, -\frac{\log \pi_{\theta} (y_w | x)}{|y_w|} - \delta
\bigg)
}$ \\
& $\beta \in [2.5, 5, 10]$, $\gamma \in [0.1, 0.5]$, $\delta \in [1, 2, 4, 6, 8]$
\\ \hline
\thickhline
\multirow{2}{*}{CPO} &
$- \log \sigma \Big(\frac{\beta} {|y_w|} \log \pi_{\theta} (y_w| x) -  \frac{\beta} {|y_l|} \log \pi_{\theta} (y_l| x) - \gamma \Big) - \frac{\lambda} {|y_w|}  \log \pi_{\theta} (y_w | x)$ \\
& $\beta \in [2.5, 5, 10]$, $\gamma \in [0.1, 0.5]$, $\lambda \in [0.1, 0.2, 0.5]$
\\ \hline
\multirow{2}{*}{DPOP} &
$- \log \sigma \Big(\beta \log \big(\frac{\pi_{\theta}(y_w|x)} {\pi_{\rm ref}(y_w|x)} \big) -\beta \log \big(\frac{\pi_{\theta}(y_l|x)} {\pi_{\rm ref}(y_l|x)}  \big) - \lambda \cdot \max \big(0, \log \big(\frac{\pi_{\rm ref}(y_w|x)} {\pi_{\theta}(y_w|x)} \big)\big) \Big)$ \\
& $\beta \in [0.01, 0.1]$, $\lambda \in [0.1, 0.2, 0.5]$
\\
 \thickhline
\end{tabular}
\caption{
Preference optimization objective functions and hyperparameter choices.
}
\label{table: preference optimization baseline algorithms implementation details}
\end{table*}

For general training hyperparameters,
we fix a batch size of $128$ for all training tasks,
and a cosine learning rate schedule with $10\%$ warmup
steps for $1$ epoch.
Preference optimization algorithms are extremely sensitive to learning rates, espectially for non-regularized implementations, such as DPO, IPO and SimPO.
Therefore, we search for the optimal learning rate from $1e^{-6}$ to $8e^{-6}$ with an increment of $1e^{-6}$.

For decoding hyperparameters, we fix a temperature of $0.6$, top-p as $0.9$, maximum token length as $2048$ for all evaluation tasks.

\subsection{Evaluation Details}
\label{sec: evaluation details}
We primarily assess our models using two of the most popular open-ended instruction-following benchmarks:
AlpacaEval $2.0$ \citep{dubois2024length} and Arena-Hard \citep{li2024live}.
AlpacaEval $2.0$ consists of $805$ questions from $5$ datasets,
Arena-Hard incorporats 500 well-defined technical problem-solving queries.

\paragraph{Model judge:}
Due to the limited access to GPT-4, we consider \href{https://huggingface.co/prometheus-eval/prometheus-bgb-8x7b-v2.0}{Mixtral-8x7B-Instruct } as the model judge \citep{kim2024prometheus} as a model judge.
This is a powerful evaluator LLM that closely mirrors human and GPT-4 judgements.
The following provides the input prompt used for model judge to compare two candidates.

\begin{lstlisting}
You are a helpful assistant, that ranks models by the quality of their answers.
Act as an impartial judge and evaluate the quality of the responses provided by two AI assistants to the user question displayed below.
The length of the response generated by each assistant is not a criterion for evaluation.
Your evaluation should consider correctness, helpfulness, completeness, and clarity of the responses.
Remember not to allow the length of the responses to influence your evaluation.
You will be given the question within <question> tags,
assistant A's answer within <assistant_a> tags,
and assistant B's answer within <assistant_b> tags.
Your job is to evaluate whether assistant A's answer or assistant B's answer is better.
Avoid any position biases and ensure that the order in which the responses are presented does not
influence your decision. Be as objective as possible.
After providing your explanation, output your final verdict within <verdict> tags strictly following this format:
<verdict>A</verdict> if assistant A is better, <verdict>B</verdict> if assistant B is better, and <verdict>tie</verdict> for a tie.
You must provide your final verdict with the format <verdict>xxx</verdict> once in your response!!!

<question>
{question}
</question>

<assistant_a>
{response_a}
</assistant_a>

<assistant_b>
{response_b}
</assistant_b>
\end{lstlisting}

\section{Extensive Experimental Results}
\label{sec: extensive experimental results}
In this section, we provide extensive numerical experimental results.

\subsection{Preference Data Construction via \methodB}
\begin{figure*}[t]
\centering
\begin{minipage}{.5\columnwidth}
\centering
    \includegraphics[width = \linewidth]{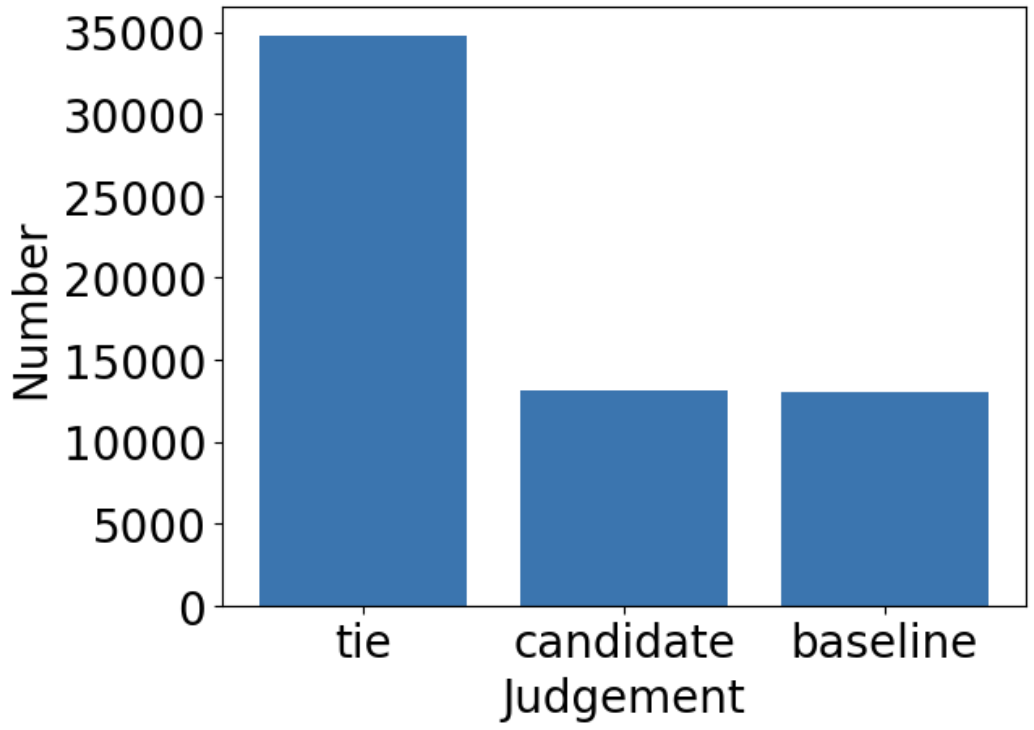}
    (a1): Iteration-$1$
\end{minipage}
\begin{minipage}{.5\columnwidth}
\centering
    \includegraphics[width = \linewidth]{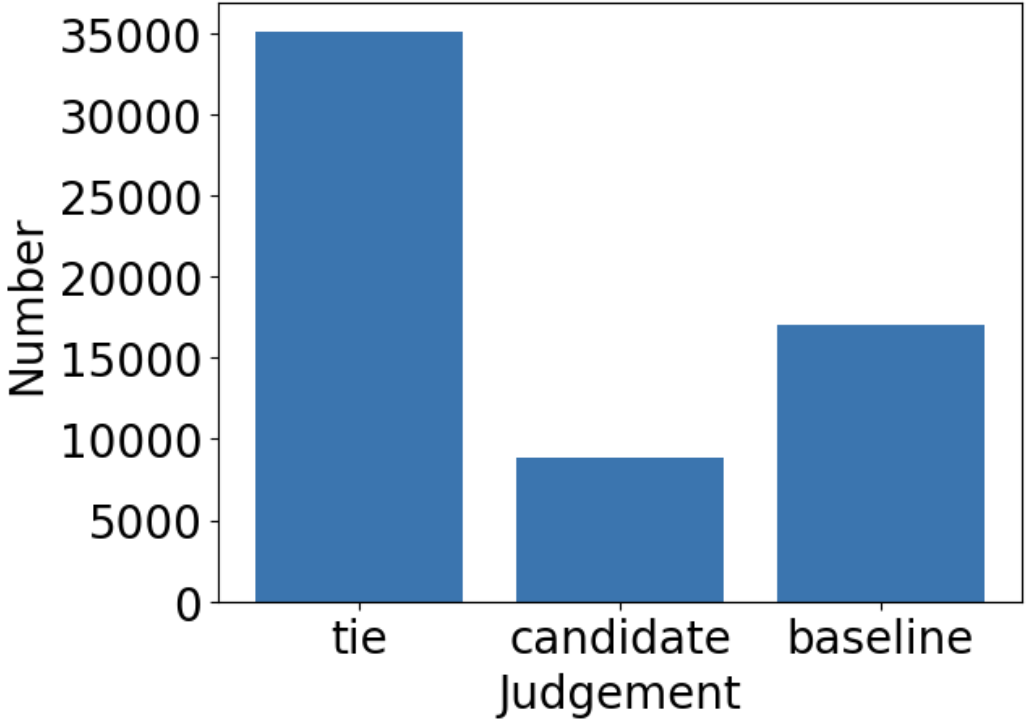}
    (b1): Iteration-$2$
\end{minipage}
\begin{minipage}{.5\columnwidth}
\centering
    \includegraphics[width = \linewidth]{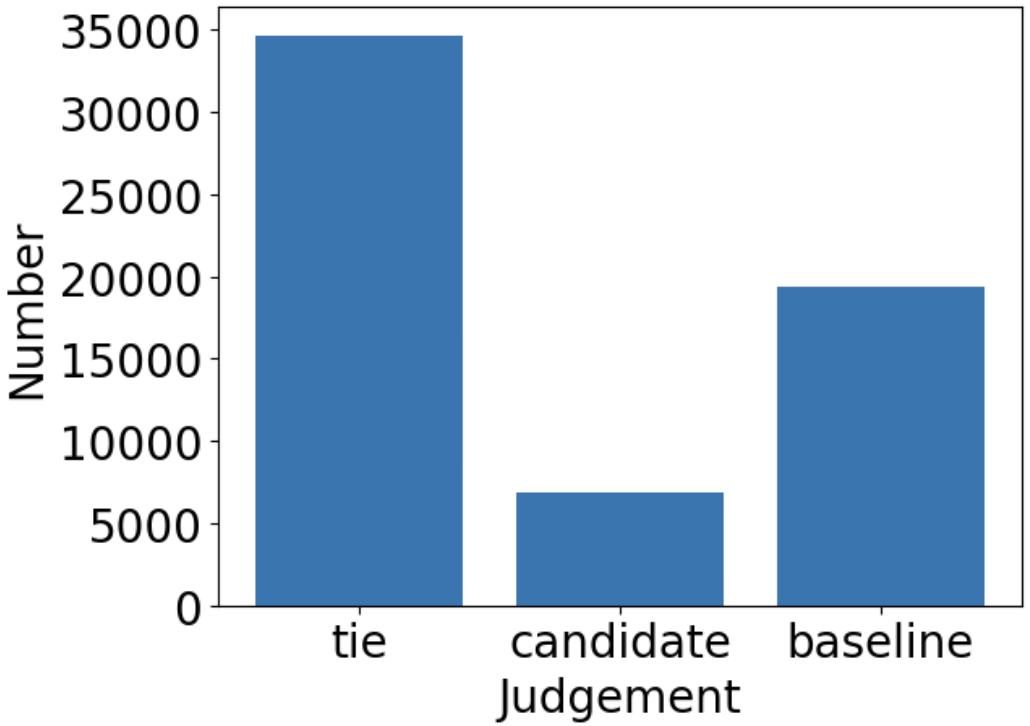}
    (c1): Iteration-$3$
\end{minipage}
\begin{minipage}{.5\columnwidth}
\centering
    \includegraphics[width = \linewidth]{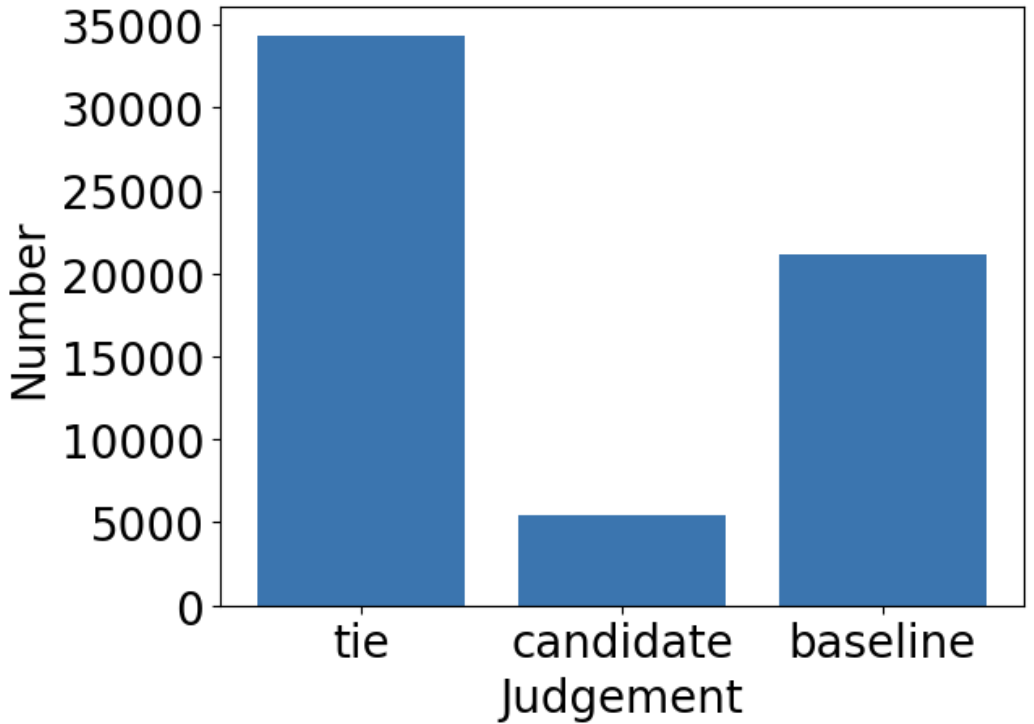}
    (d1): Iteration-$4$
\end{minipage}

\begin{minipage}{.5\columnwidth}
\centering
    \includegraphics[width = \linewidth]{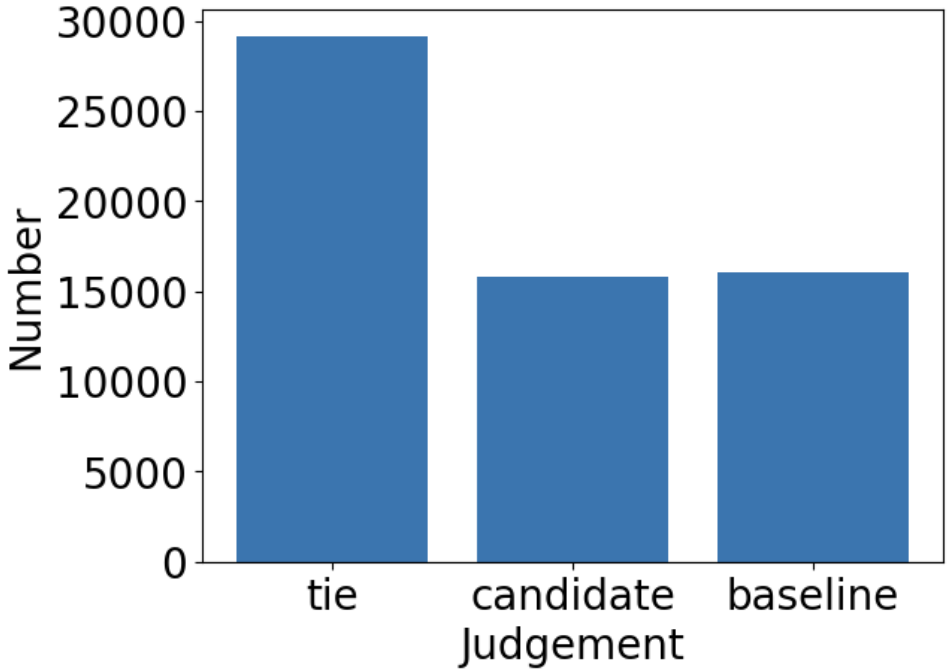}
    (a2): Iteration-$1$
\end{minipage}
\begin{minipage}{.5\columnwidth}
\centering
    \includegraphics[width = \linewidth]{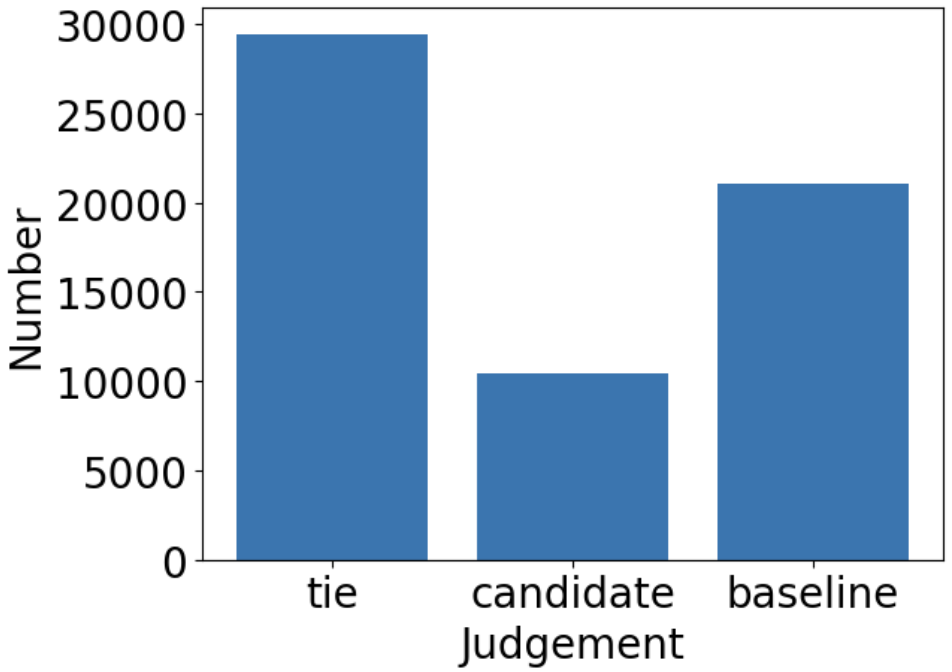}
    (b2): Iteration-$2$
\end{minipage}
\begin{minipage}{.5\columnwidth}
\centering
    \includegraphics[width = \linewidth]{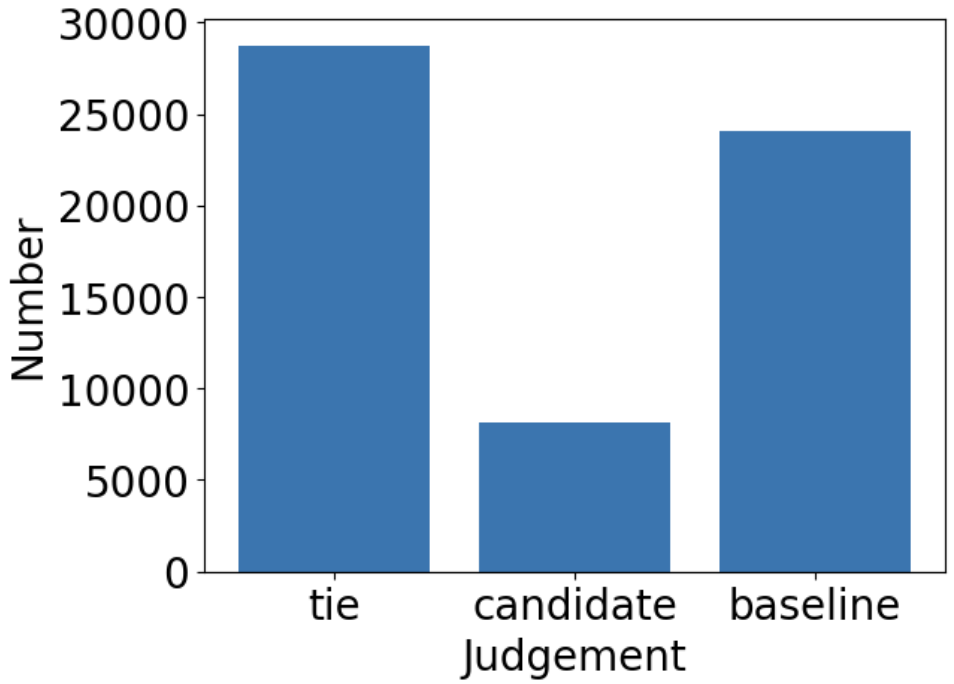}
    (c2): Iteration-$3$
\end{minipage}
\begin{minipage}{.5\columnwidth}
\centering
    \includegraphics[width = \linewidth]{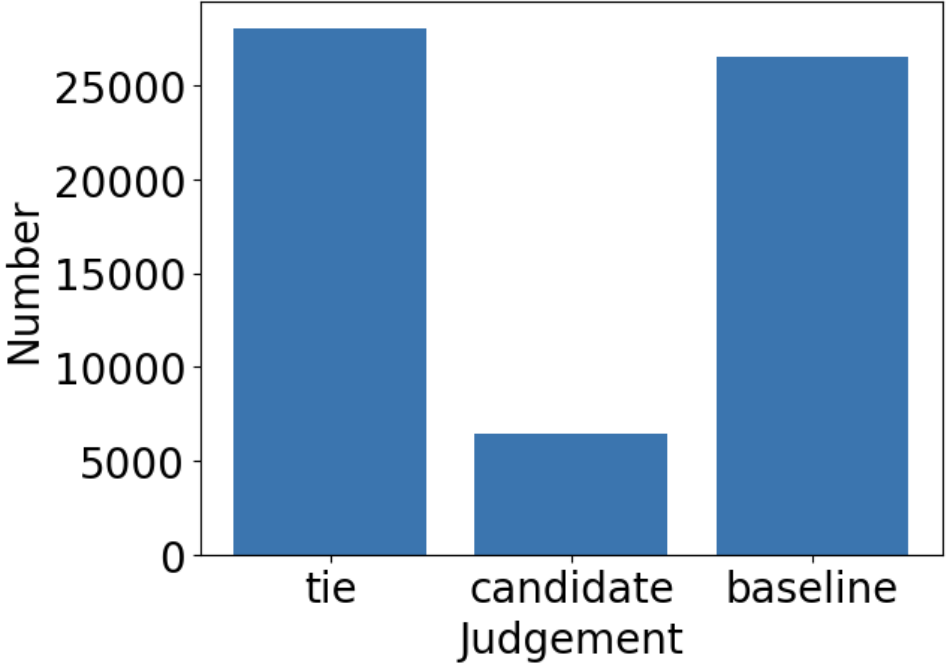}
    (d2): Iteration-$4$
\end{minipage}
\caption{
Statistics of \methodB.
For \methodB(Llama70B) with Llama-3.1-Instruct as base model: (a1), (b1), (c1) and (d1) present the statistics of preference comparisons at all $4$ iterations.
For \methodB(Llama70B) with Mistral-Instruct as base model: (a2), (b2), (c2) and (d2) present the statistics of preference comparisons at all $4$ iterations.
}
\label{fig: ipr iteration statistics}
\end{figure*}

We create a preference dataset by using completions generated from the base model, which helps reduce the gap between the base model's outputs and the preference optimization process. 
For each input prompt, we generate five candidate completions and use our proposed \methodB~ method to select the most preferred one. 
Figure \ref{fig: ipr iteration statistics} shows the statistics for \methodB(Llama70B) (using Llama-3.1-70B as the preference judge).

Each comparison can result in one of three outcomes: Tie, Candidate, or Baseline. Since all candidate completions come from the same distribution (the base model), a large number of Ties occur in each iteration. In cases of a Tie, we always select the baseline completion as the winner. If all four iterations result in Ties, we choose the first candidate completion. This preferred completion is still of high quality because it is at least as good as the other candidates.

\label{sec: preference data generation via ipr}

\subsection{Preference Optimization Regularization}
\label{appendix: preference optimization regularizations}

\begin{figure*}[t]
\centering
\begin{minipage}{.65\columnwidth}
\centering
    \includegraphics[width = \linewidth]{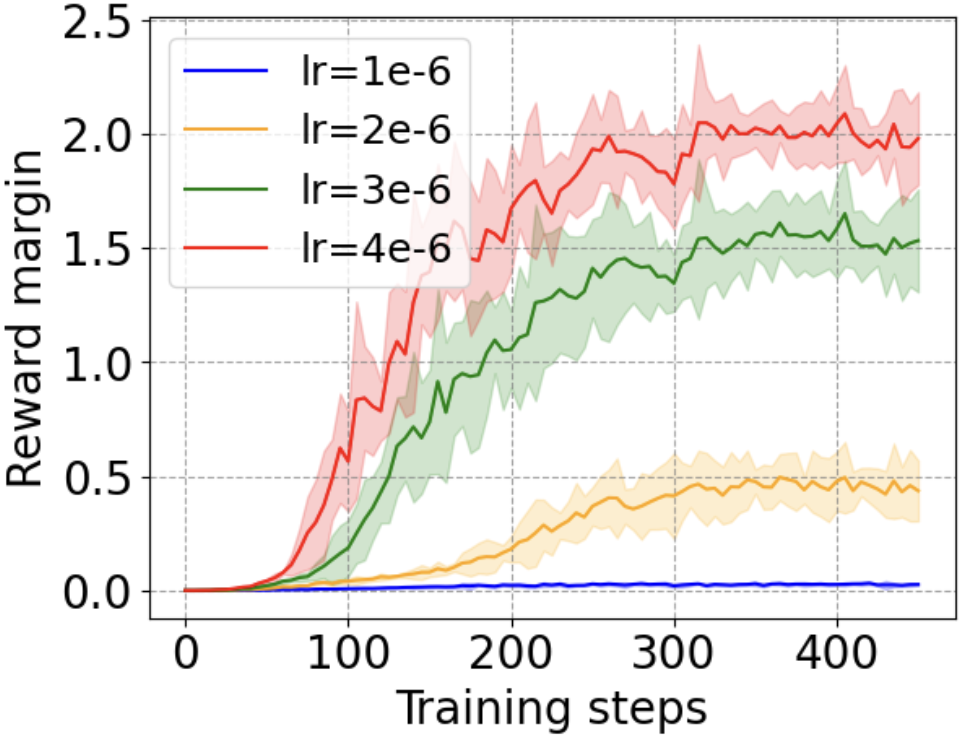}
    (a1): Reward margin
\end{minipage}
\begin{minipage}{.65\columnwidth}
\centering
    \includegraphics[width = \linewidth]{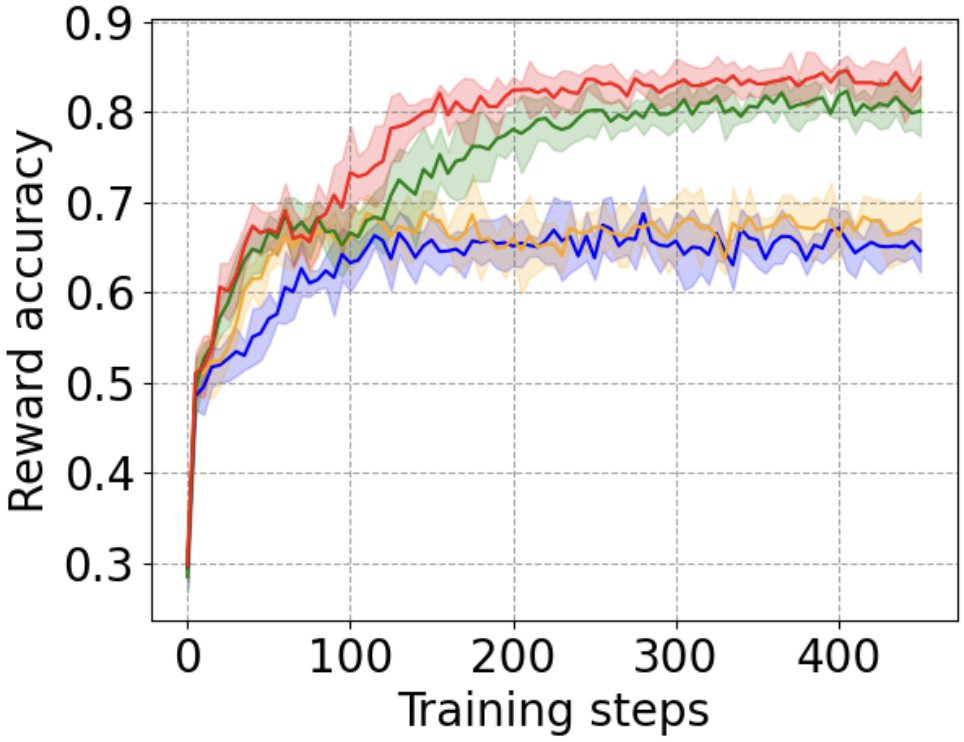}
    (b1): Reward accuracy
\end{minipage}
\begin{minipage}{.65\columnwidth}
\centering
    \includegraphics[width = \linewidth]{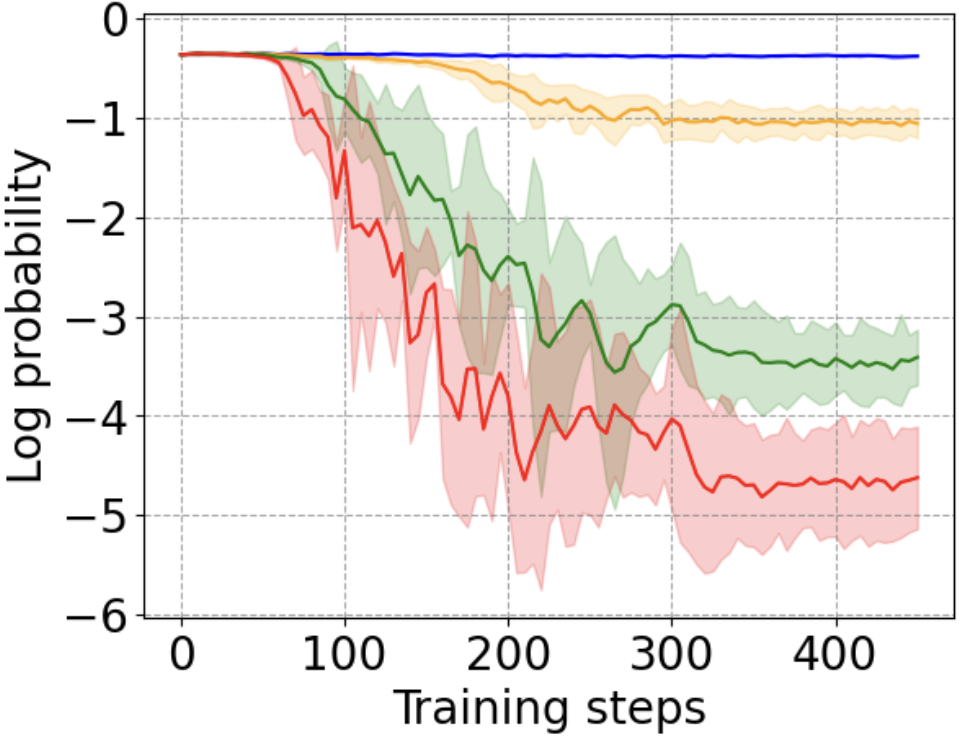}
    (c1): $\pi_{\theta}(y_w | x)$
\end{minipage}

\begin{minipage}{.65\columnwidth}
\centering
    \includegraphics[width = \linewidth]{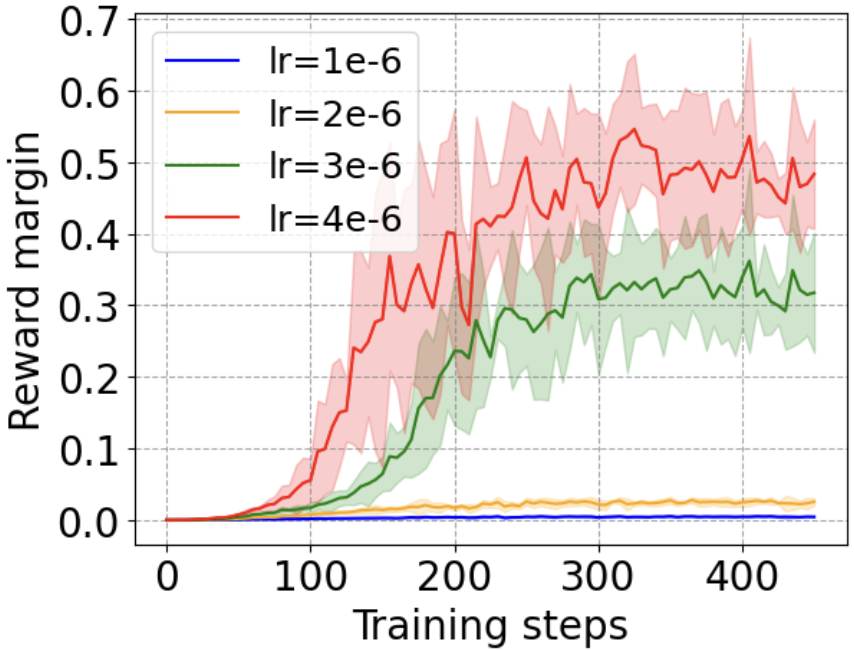}
    (a2): Reward margin
\end{minipage}
\begin{minipage}{.65\columnwidth}
\centering
    \includegraphics[width = \linewidth]{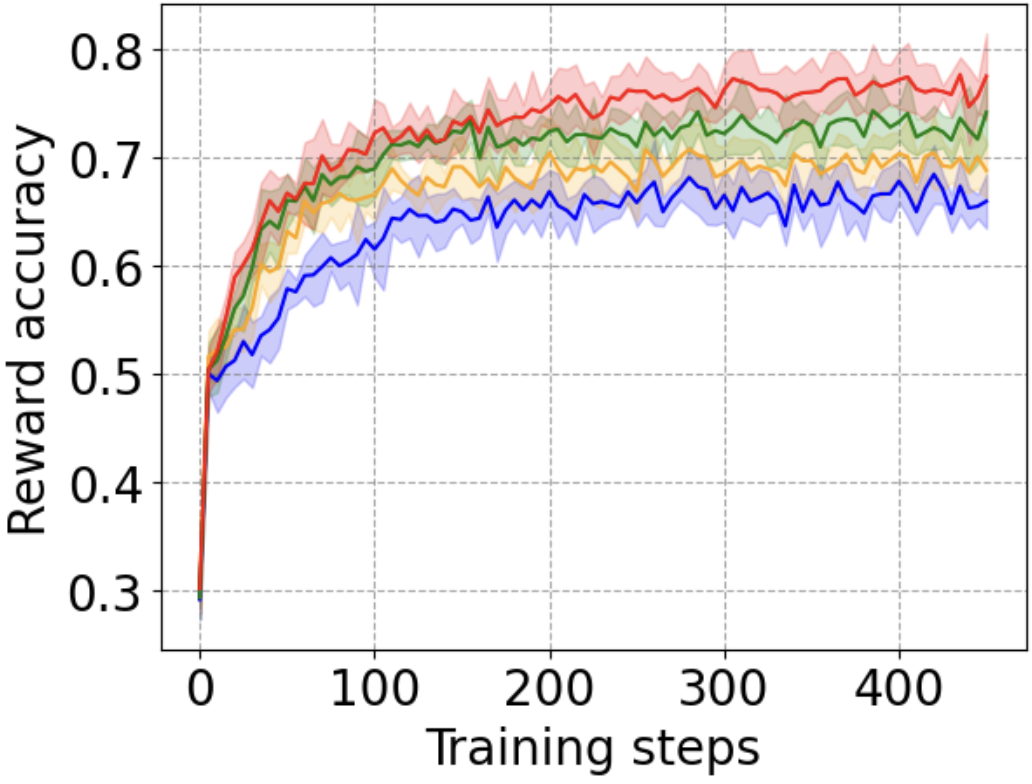}
    (b2): Reward accuracy
\end{minipage}
\begin{minipage}{.68\columnwidth}
\centering
    \includegraphics[width = \linewidth]{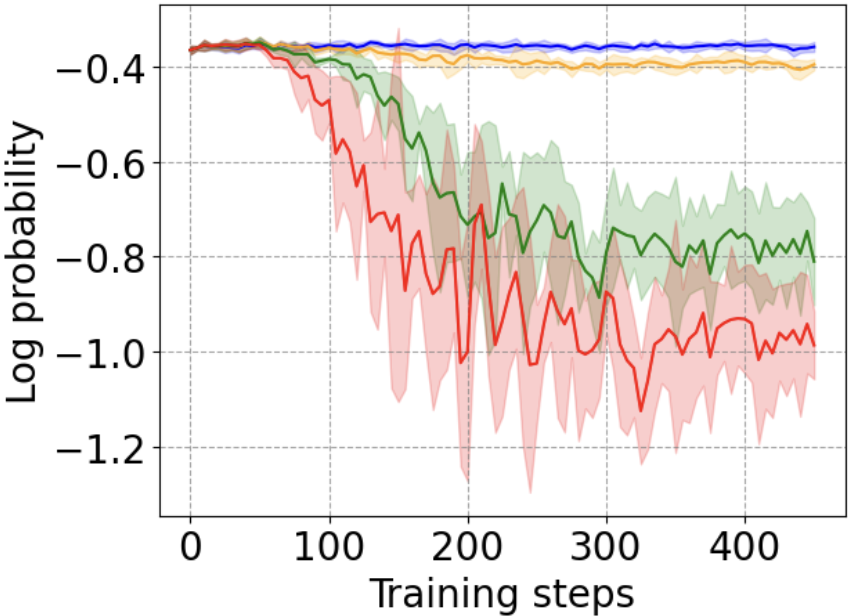}
    (c3): $\pi_{\theta}(y_w | x)$
\end{minipage}

\begin{minipage}{.65\columnwidth}
\centering
    \includegraphics[width = \linewidth]{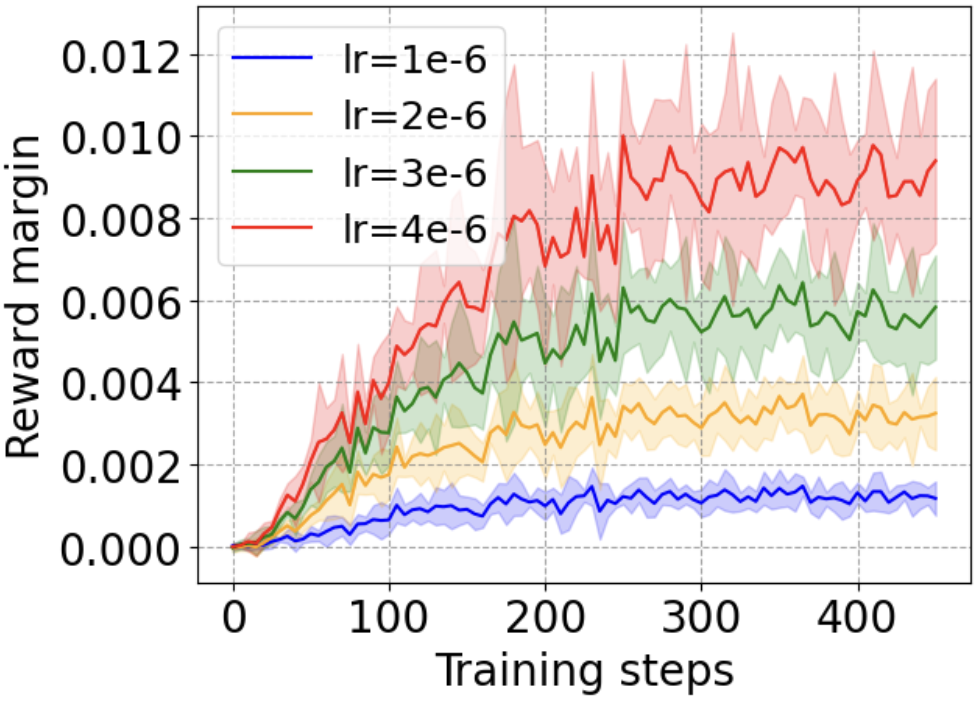}
    (a3): Reward margin
\end{minipage}
\begin{minipage}{.65\columnwidth}
\centering
    \includegraphics[width = \linewidth]{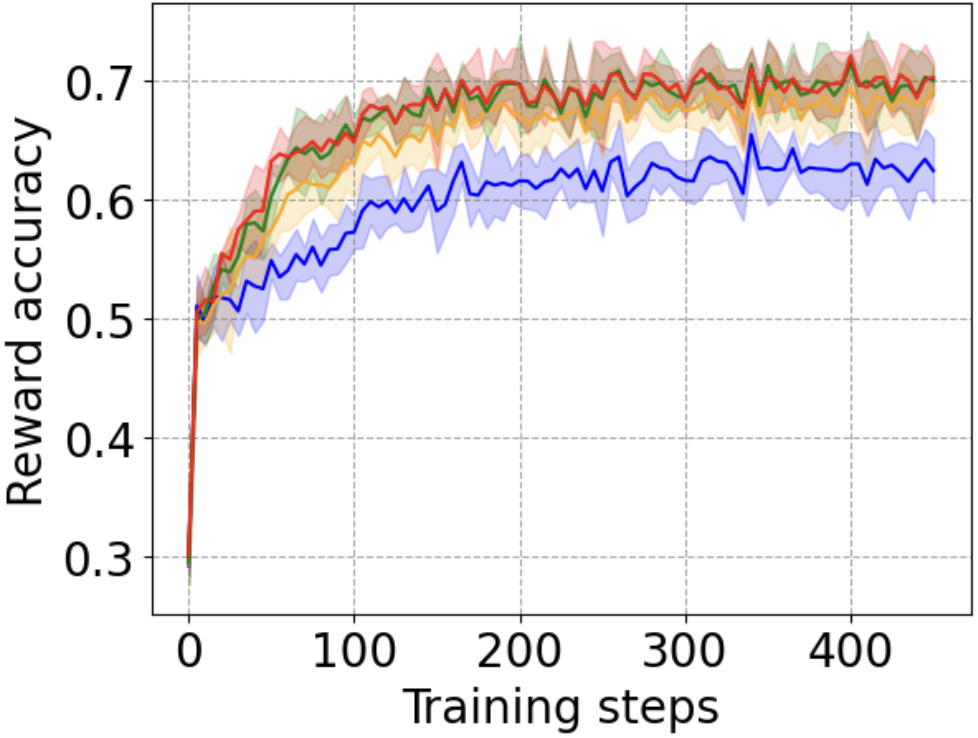}
    (b3): Reward accuracy
\end{minipage}
\begin{minipage}{.7\columnwidth}
\centering
    \includegraphics[width = \linewidth]{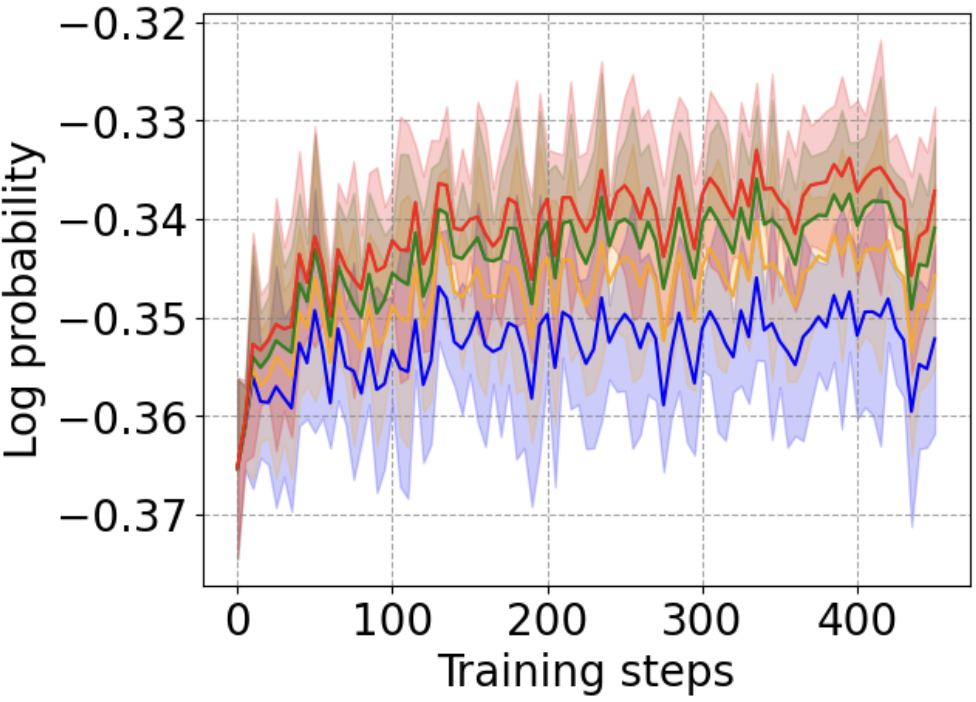}
    (c3): $\pi_{\theta}(y_w | x)$
\end{minipage}
\caption{
Training progress for DPO and DPOP. 
(a1), (b1), and (c1) display the reward margin, reward accuracy, and log-likelihood of predicting preferred completions for DPO, respectively. 
(a2), (b2), and (c2) present the same metrics for DPOP with $\lambda=0.5$, while (a3), (b3), and (c3) show the training progresses for DPOP with $\lambda=5$. 
Each configuration is evaluated using four different learning rates: $1e-6$, $2e-6$, $3e-6$, and $4e-6$.
}
\label{fig: dpo log prob reward margin different learning rates}
\end{figure*}

\begin{figure*}[t]
\centering
\begin{minipage}{.65\columnwidth}
\centering
    \includegraphics[width = \linewidth]{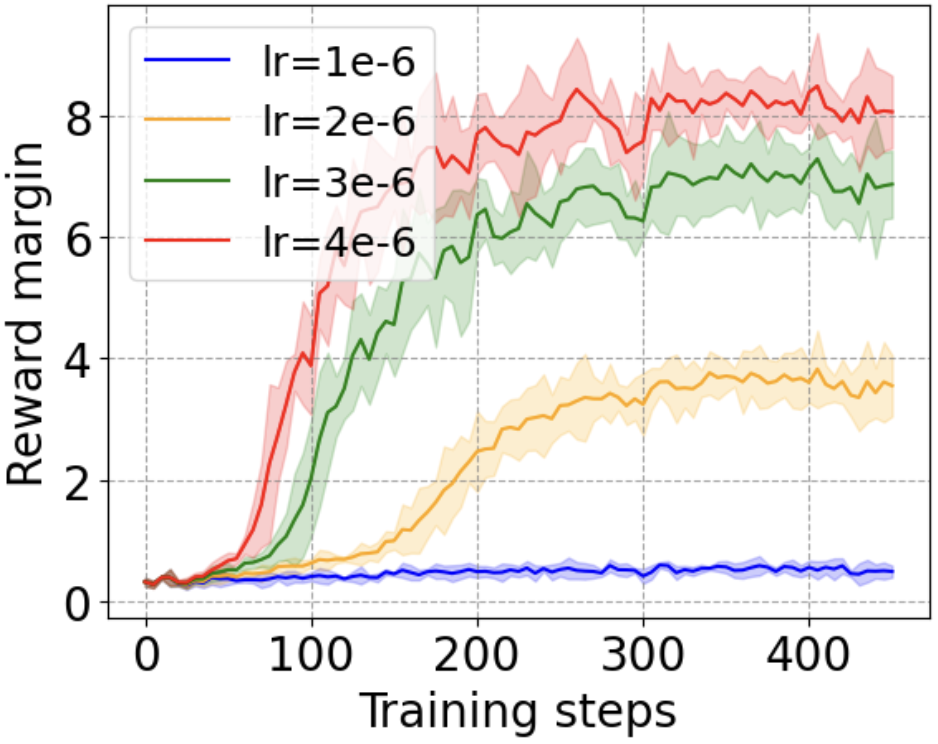}
    (a1): $r(x, y_w) - r(x, y_l)$
\end{minipage}
\begin{minipage}{.7\columnwidth}
\centering
    \includegraphics[width = \linewidth]{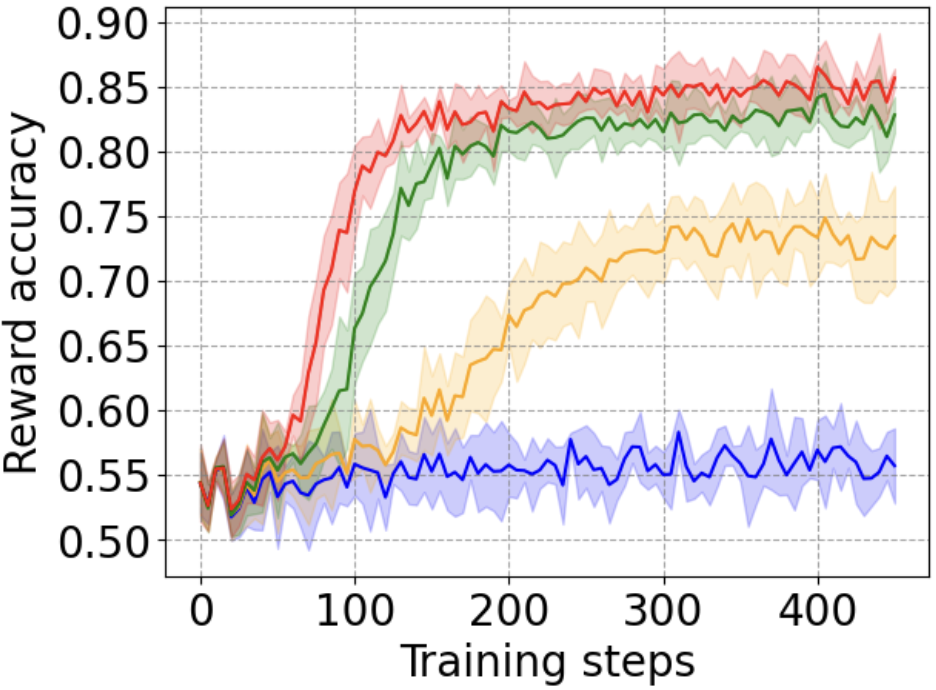}
    (b1): Reward accuracy
\end{minipage}
\begin{minipage}{.7\columnwidth}
\centering
    \includegraphics[width = \linewidth]{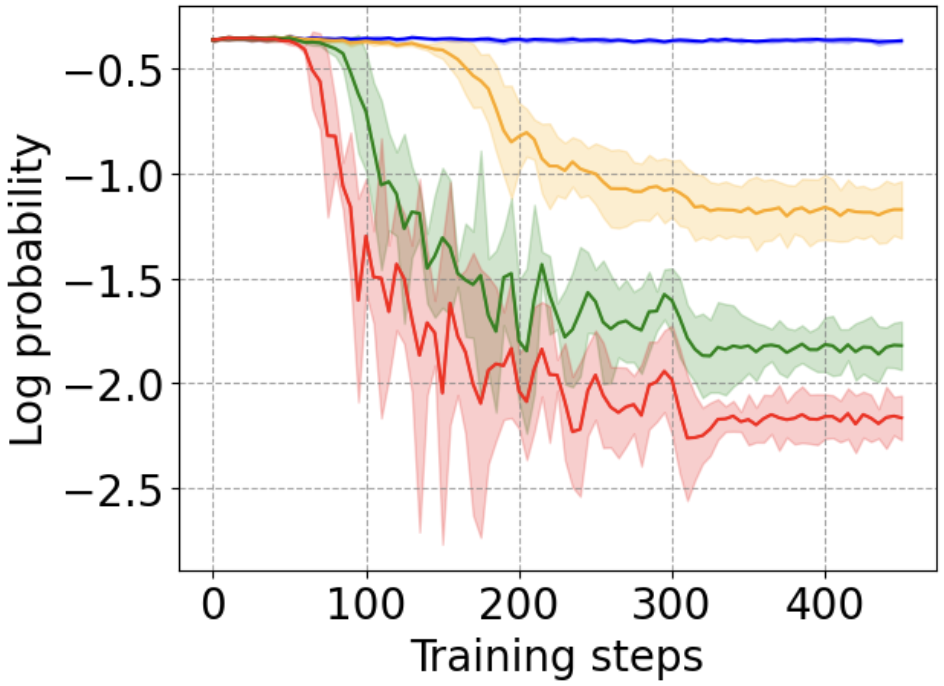}
    (c1): $\pi_{\theta}(y_w | x)$
\end{minipage}

\begin{minipage}{.65\columnwidth}
\centering
    \includegraphics[width = \linewidth]{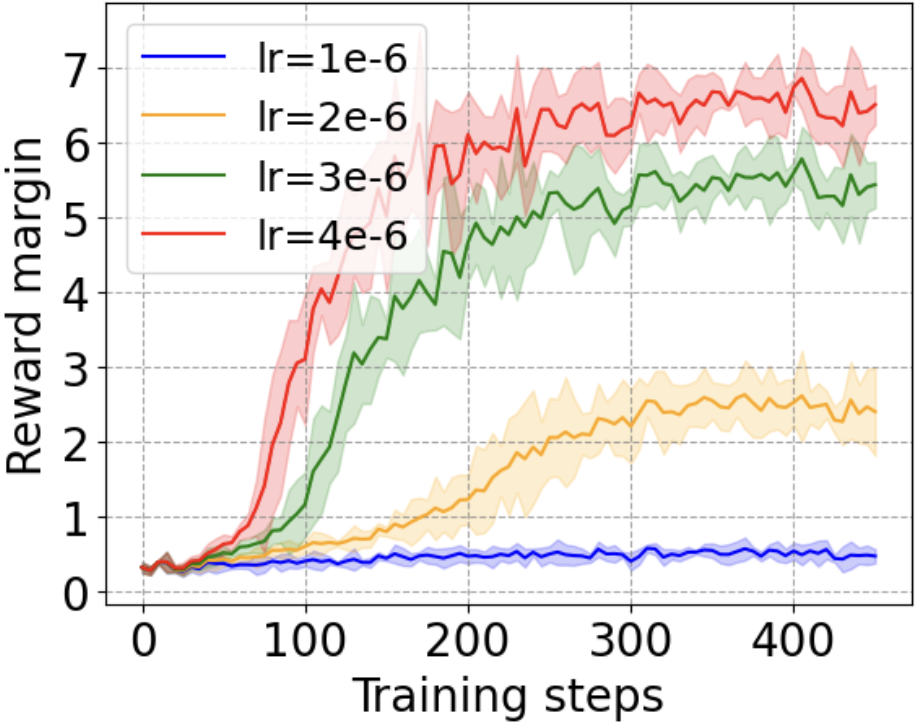}
    (a2): $r(x, y_w) - r(x, y_l)$
\end{minipage}
\begin{minipage}{.7\columnwidth}
\centering
    \includegraphics[width = \linewidth]{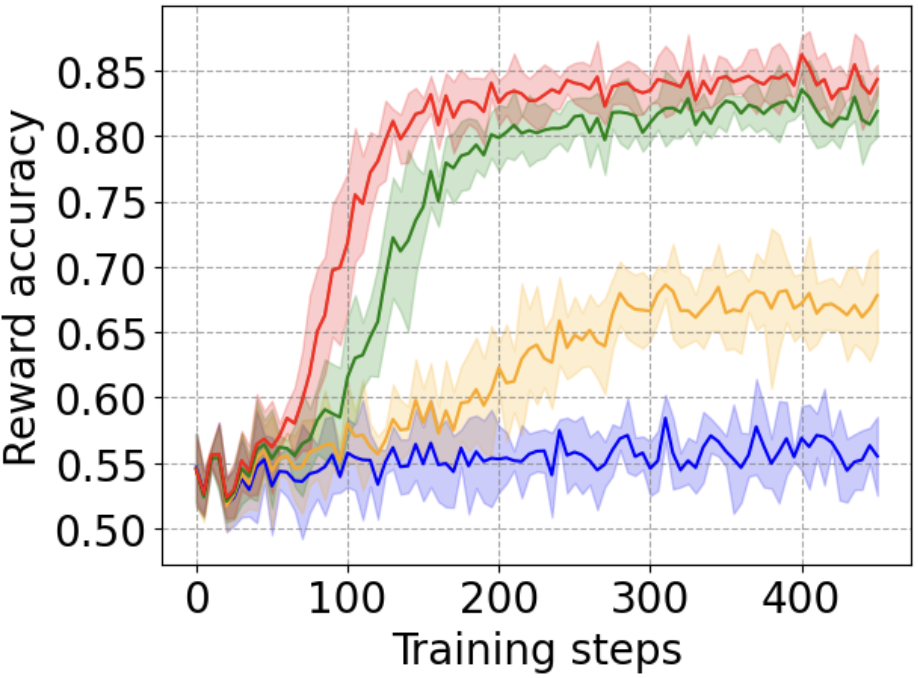}
    (b2): Reward accuracy
\end{minipage}
\begin{minipage}{.7\columnwidth}
\centering
    \includegraphics[width = \linewidth]{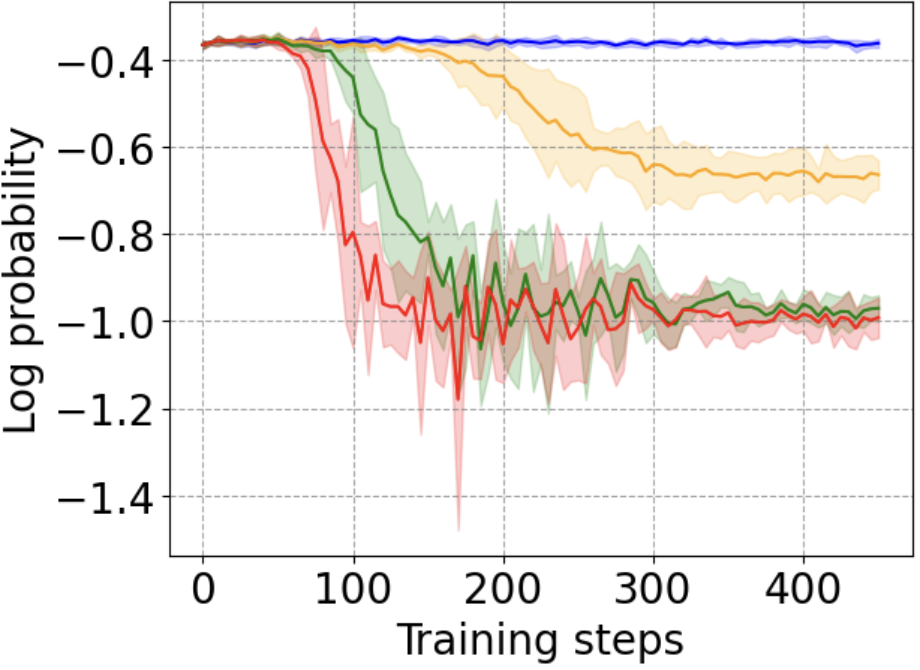}
    (c2): $\pi_{\theta}(y_w | x)$
\end{minipage}

\begin{minipage}{.65\columnwidth}
\centering
    \includegraphics[width = \linewidth]{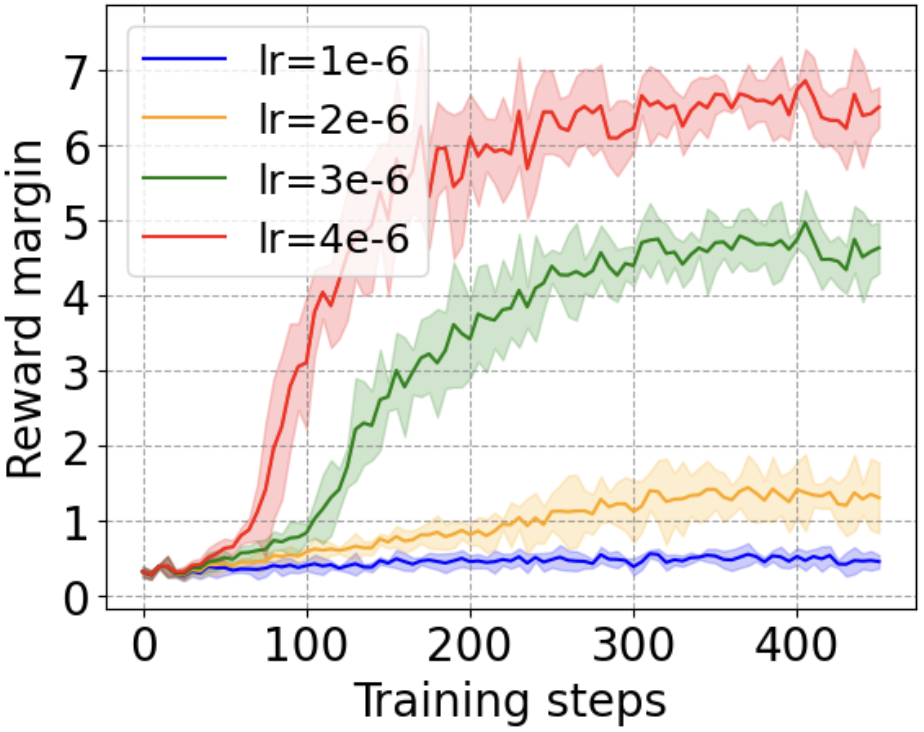}
    (a3): $r(x, y_w) - r(x, y_l)$
\end{minipage}
\begin{minipage}{.7\columnwidth}
\centering
    \includegraphics[width = \linewidth]{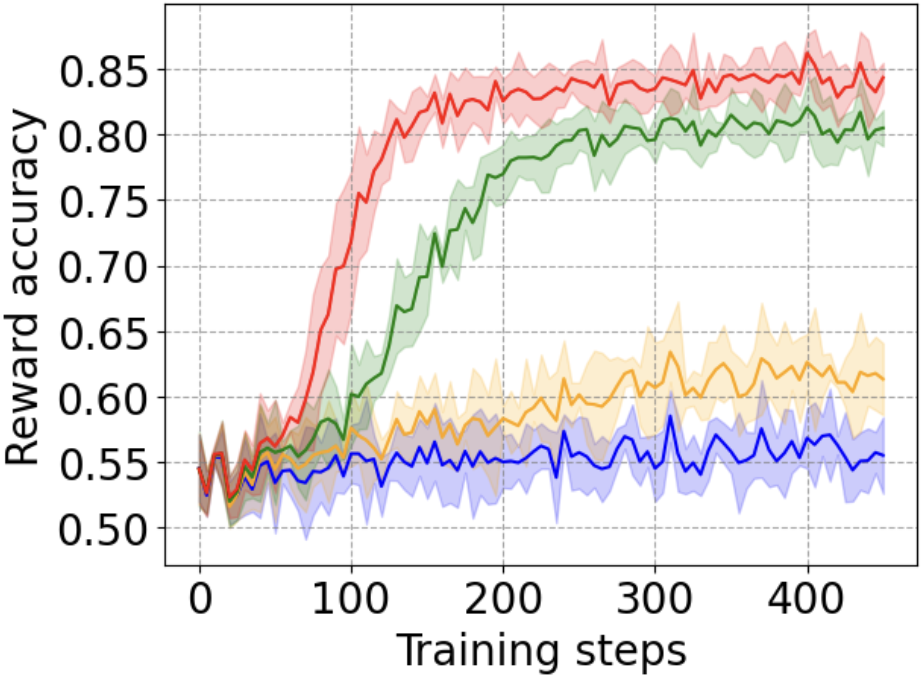}
    (b3): Reward accuracy
\end{minipage}
\begin{minipage}{.7\columnwidth}
\centering
    \includegraphics[width = \linewidth]{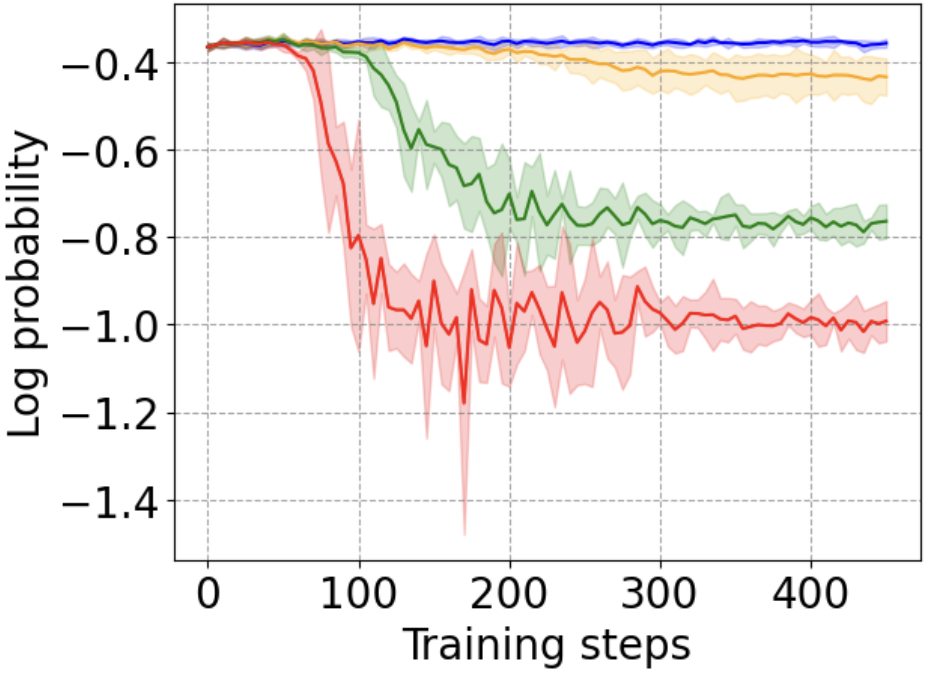}
    (c3): $\pi_{\theta}(y_w | x)$
\end{minipage}
\caption{
Training progress for SimPO and CPO. 
(a1), (b1), and (c1) display the reward margin, reward accuracy, and log-likelihood of predicting preferred completions for SimPO, respectively. 
(a2), (b2), and (c2) present the same metrics for CPO with $\lambda=0.5$, while (a3), (b3), and (c3) show the training progresses for DPOP with $\lambda=1$. 
Each configuration is evaluated using four different learning rates: $1e-6$, $2e-6$, $3e-6$, and $4e-6$. 
}
\label{fig: simpo log prob reward margin different learning rates}
\end{figure*}

\paragraph{DPO versus DPOP results:}
Here we provide extensive results to showcase the failure mode in preference optimization.
Figure \ref{fig: dpo log prob reward margin different learning rates} shows the training progresses for DPO and DPOP.
In Figure \ref{fig: dpo log prob reward margin different learning rates} (a1), (b1) and (c1),
as both reward margin and reward accuracy increases, DPO leads to a reduction on the log-likelihood of predicting preferred completions.
When the supervised next-word prediction regularization is added by setting $\lambda=0.5$ in DPOP, 
in Figure \ref{fig: dpo log prob reward margin different learning rates} (a2), (b2) and (c2),
the issue of reducing log-likelihood of predicting preferred completion is alleviated, 
however,
the reward accuracy is lower compared to DPO in Figure \ref{fig: dpo log prob reward margin different learning rates} (a2).
When the regularization effect is stronger with a larger $\lambda=5$,
the log-likelihood of predicting preferred completion is non-decreasing through the whole training progress.
However,
the reward accuracy is considerably lower compared to DPO in Figure \ref{fig: dpo log prob reward margin different learning rates} (b1).

\paragraph{SimPO Versus CPO results:}
~Figure \ref{fig: simpo log prob reward margin different learning rates} illustrates the training progress of SimPO and CPO (SimPO with regularization). 
In Figure \ref{fig: simpo log prob reward margin different learning rates} (a1), (b1), and (c1), as both reward margin and reward accuracy increase, SimPO results in a reduction in the log-likelihood of predicting preferred completions. 
However, when supervised next-word prediction regularization is introduced by setting $\lambda=0.5$ in CPO, as shown in Figure \ref{fig: simpo log prob reward margin different learning rates} (a2), (b2), and (c2), this issue is alleviated. 
Nonetheless, the reward accuracy in CPO is lower compared to SimPO. 
When the regularization is made stronger with $\lambda=1$, the reward accuracy decreases significantly, as seen in Figure \ref{fig: simpo log prob reward margin different learning rates} (b1) compared to SimPO.

\section{Efficient Preference Data Generation}
\label{sec: efficient preference data generation}
\paragraph{An early stopping criterion.}
Given consideration of computational efficiency, the goal is to explore the preferred completion while minimizing the number of comparison signals, which can be computationally expensive (such as using an LLm judge).
The threshold-based stopping criterion aims to stop exploration when there is sufficient evidence that one completion is preferred over all others \citep{bubeck2009pure, zoghi2014relative}.
We define this criterion using prior estimations for all possible pairwise comparisons.
Recall that each comparison signal has $3$ possible outcomes, baseline wins, candidate wins and a tie.
In the exhaustive search process, we select the outcome from the first non-tie comparison as the overall preferred completion.

This approach is motivated by the online preference optimization setting, where candidate completions are generated by sampling from the same distribution in the target LLM and there is a high probability that many comparisons will result in ties. 
Therefore, by selecting the first non-tie outcome, the process can be stopped early, avoiding unnecessary comparisons.

\section{Related Works}
\label{sec: appendix related works}
In this section, we first outline DPO and its variants,
then we discuss the training instability issue associated to these preference optimization algorithms and existing solutions.

\paragraph{DPO and Its Variants.}
~Since the introduction of DPO \citep{rafailov2024direct}, several algorithms have emerged to further refine preference optimization. 
SimPO (Simple Preference Optimization) introduces length regularization on the log-probabilities of both preferred and dispreferred completions, eliminating the need for a reference model, as required in DPO \citep{meng2024simpo}. 
This method improves model alignment while reducing computational demands.
IPO (Identity Preference Optimization) addresses the shortcomings of Bradley-Terry preference modeling in cases where preference data are highly deterministic, when the preferred completion is almost always better to the dispreferred one. 
In such cases, the KL-divergence regularization becomes ineffective.
IPO resolves this by replacing the logistic loss with a squared loss and incorporating a margin, providing a more theoretically sound approach \citep{azar2024general}.
Other notable algorithms include SLIC (sequence likelihood calibration), which applies a ranking calibration loss between preferred and dispreferred completions \citep{zhao2023slic}, 
RPO (Regularized preference optimization), emphasizing the role of length regularization \citep{park2024disentangling}, and $\beta$-PO, which dynamically adjusts the $\beta$ hyperparameter at the batch level \citep{wu2024beta}. 
TRPO (Trust Region Preference Optimization) updates the reference policy during training, improving stability \citep{gorbatovski2024learn}, iterative preference learning iteratively refine the target LLM based on preference data, progressively improving performance \citep{xiong2024iterative, kim2024sdpo}.
In this work, we show that the performance of existing preference optimization algorithms can be further improved with higher quality preference data.

\paragraph{Supervised Next-Word Prediction Regularization Improves Training Stability.}
~DPO models the relative probability of selecting one completion over another using pairs of preferred and non-preferred data. 
However, the standard DPO loss may inadvertently reduce the model's likelihood of producing the preferred completion, as long as the relative probability between the preferred and non-preferred completions increases \citep{feng2024towards}. 
This can result in a failure mode during DPO training \citep{pal2024smaug}. 
To address this, various forms of supervised next-word prediction regularization have been proposed to improve training stability. 
For example, SLIC adds a term to maximize log-likelihoods on certain reference completions \citep{zhao2023slic}, while CPO (Contrastive Preference Optimization) applies a behavior cloning regularizer that specifically optimizes the preferred completions \citep{hejnacontrastive,xucontrastive}.
Additionally, DPOP introduces a hinge loss on the log-ratio between the reference and target models \citep{pal2024smaug}. 
Despite the improvements in training stability, our analysis indicates that regularized preference optimization often results in worse performance compared to non-regularized approaches.

\end{document}